\documentclass[11pt,letterpaper]{article}
\usepackage[letterpaper,hmargin=0.88in,vmargin=0.76in]{geometry}
\usepackage{times}
\usepackage[round,authoryear]{natbib}
\usepackage{tabularx}

\usepackage{amsmath,amsfonts,bm}

\def\eqref#1{equation~\ref{#1}}

\def\1{\bm{1}}

\DeclareMathAlphabet{\mathsfit}{\encodingdefault}{\sfdefault}{m}{sl}
\SetMathAlphabet{\mathsfit}{bold}{\encodingdefault}{\sfdefault}{bx}{n}

\usepackage{graphicx}
\usepackage{amsmath,amssymb,amsthm}
\usepackage{booktabs}
\usepackage{multirow}
\usepackage{makecell}
\usepackage{xcolor}
\usepackage{subcaption}
\usepackage{float}
\usepackage{placeins}
\usepackage{hyperref}
\hypersetup{hidelinks,pdfauthor={Pengjie Zheng, Mengwei Xie, Huan Yang, Si Wu, Pengfei Hu, Feng Li},pdftitle={AVERT: Average-Depth Variable Execution via Request-Local Token Routing}}
\usepackage{url}
\usepackage{microtype}
\floatstyle{ruled}
\newfloat{algorithm}{tbp}{loa}
\floatname{algorithm}{Algorithm}
\floatstyle{plain}

\newcommand{\method}{AVERT}
\newcommand{\omegaT}{\Omega}

\title{\method{}: Average-Depth Variable Execution via Request-Local Token Routing}
\author{%
Pengjie Zheng\textsuperscript{1}\enspace Mengwei Xie\textsuperscript{1}\enspace Huan Yang\textsuperscript{2}\enspace Si Wu\textsuperscript{1}\enspace Pengfei Hu\textsuperscript{1}\enspace Feng Li\textsuperscript{1}\\
{\small \textsuperscript{1}School of Computer Science and Technology, Shandong University, Qingdao 266237, China}\\
{\small \textsuperscript{2}College of Computer Science and Technology, Qingdao University, Qingdao 266071, China}\\
{\footnotesize \href{mailto:pjzheng@mail.sdu.edu.cn}{\nolinkurl{pjzheng@mail.sdu.edu.cn}}\quad \href{mailto:xmw_sdu@126.com}{\nolinkurl{xmw_sdu@126.com}}\quad \href{mailto:yanghuan@qdu.edu.cn}{\nolinkurl{yanghuan@qdu.edu.cn}}}\\
{\footnotesize \href{mailto:kobesi@sdu.edu.cn}{\nolinkurl{kobesi@sdu.edu.cn}}\quad \href{mailto:phu@sdu.edu.cn}{\nolinkurl{phu@sdu.edu.cn}}\quad \href{mailto:fli@sdu.edu.cn}{\nolinkurl{fli@sdu.edu.cn}}}}
\date{}

\begin{document}
\maketitle

\begin{abstract}
Layer skipping can reduce large language model inference costs, but a request budget does not specify how to allocate layer execution across generated tokens. We introduce \method{}, which allows tokens to execute different numbers of layers while matching each request's mean executed depth to a continuous budget. At each routed layer, a router reads the current token state and its request's budget and decides whether to execute the Transformer block. Tokens that skip the block pass through a lightweight residual adapter and write no key/value (K/V) entry at that layer. A single checkpoint serves requests with different budgets in the same batch. Across three LLaMA backbones, six quality metrics, and 13 budgets, \method{} is best or tied for best in 206 of 234 comparisons. Across single- and mixed-budget batches, the mean absolute per-request budget error is at most 1.15 percentage points. When grouping requests by budget leaves batches underfilled, mixing budgets reduces decode time per output token (TPOT) by 4.3--21.2\%.
\end{abstract}

\section{Introduction}
\label{sec:intro}
Large language model (LLM) serving systems batch requests to improve utilization~\citep{orca,vllm,sarathiServe}. A service may assign different compute budgets even to the same input, depending on its cost or latency requirements. The budget expresses how much computation to spend; the model must decide where to spend it within the response. We study this allocation through \emph{interior layer skipping}, where a token can bypass a Transformer block and still execute later blocks.

Existing methods use budgets in different ways. In D-LLM, \citet{d-llm} train a separate model for each budget and penalize each token's deviation from the target depth. In BUDDY, \citet{buddy} use each request's budget to fix the number of layers executed by every token. In GateSkip, \citet{WLW} use the budget to select a fraction of the batch's live tokens for execution at each attention or feed-forward network (FFN) sublayer. In L2A, \citet{L2A} use a continuous budget to target the combined cost of layers and attention heads averaged over a mini-batch and to control reasoning length. We study how one model can meet each request's average-depth target while allowing tokens in the same response to execute different numbers of layers.

A request with a budget of 0.6 asks \method{} to execute 60\% of its layers per routed token, on average. When a token reaches a routed layer, the router decides whether to run the block using only the token's current hidden state and its request's budget. Tokens in one response can execute different numbers of layers. At the same position in a fixed text, a token can execute a layer at a lower budget and skip that same layer at a higher budget (Figure~\ref{fig:budget-nesting}). We train one model across a range of sampled budgets. For each request, we penalize the absolute difference between its budget and the average fraction of layers its routed tokens execute.

In DistServe, \citet{distServe} allocate separate GPU resources to prefill and decoding, allowing each stage to be optimized independently. We focus on reducing decoding costs while keeping prefill at full depth. During decoding, \method{} processes requests with different budgets in the same batch. At each layer, it gathers the tokens that will execute the block and processes them together. A token that skips the block writes no key/value (K/V) entry at that layer, so later tokens from the same request attend to a shorter history there. A lightweight skip-path adapter compensates for the representation change caused by bypassing a block, helping the shared checkpoint maintain quality across budgets.

On three LLaMA backbones, six quality metrics, and 13 requested budgets, \method{} is best or tied for best in 206 of the 234 comparisons. For each backbone and budget, we average the absolute budget errors of requests assigned that budget. The largest of these averages is 0.90 percentage points on five off-grid targets and 1.15 points in mixed-budget batches. In controlled workloads, mixing budgets has comparable decode TPOT when batches separated by budget remain full; it lowers TPOT by 4.3--21.2\% when separation creates underfilled batches. Our ablation compares request-average and per-token objectives using the same architecture and decoding implementation, testing the benefit of targeting the request's mean depth rather than each token's depth.

Our contributions are (i) a formulation and training objective for matching each request's mean executed depth to a supplied target, (ii) a comparison showing that targeting request-average depth improves generation quality over a per-token depth objective at comparable average execution, and (iii) an implementation and evaluation of decoding requests with different budgets using one checkpoint, covering quality, budget tracking, cache storage, and the efficiency of batching requests with different budgets.

\section{Related Work}
\label{sec:related}
\paragraph{Budgeted conditional computation.}
Several methods condition computation on an external resource budget. In CCT, \citet{CCT} use budget control symbols to govern Transformer operations and penalize aggregate computation over a batch, allowing examples to share the available compute. In SOL, \citet{SOL} condition a policy on requested budgets and select attention sparsity, FFN-channel pruning, and activation quantization actions, matching average usage over short decode episodes. In L2A, \citet{L2A} condition layer, head, and reasoning-transition gates on a continuous resource signal. The training objective averages normalized layer and head costs over a mini-batch and separately penalizes reasoning length. In \method{}, a budget of 0.6 targets 60\% layer execution on average over a request's routed tokens. The budget error is computed separately for each request.

\paragraph{Shared layer skipping.}
Some methods apply the same layer selection to multiple tokens. In FiRST, \citet{FiRST} select which layers to execute based on the prompt and apply the same selection to all generated tokens in that request. In SkipDecode, \citet{skipdecode} assign the same execution depth to tokens at the same sequence position across a batch, with depth decreasing as generation progresses. In AdaSkip, \citet{adaskip} use offline importance estimates to select the attention and FFN sublayers to execute, with the selected set shared across requests for a given acceleration ratio. \citet{ReducingDepth} train reduced-depth subnetworks with LayerDrop. In Hierarchical Skip Decoding, \citet{hierarchicalSkip} select layers according to the current sequence length. In Unified Layer Skipping, \citet{UnifiedStrategy} select a balanced input-independent pattern for a target speedup. In DASH, \citet{MarkovDash} model input-aware layer execution states, including skipping and quantized computation. Early-exit methods produce predictions from intermediate layers~\citep{free,adaInfer}. Self-speculative methods draft tokens with reduced computation and verify them using the full model~\citep{draft,LayerskipEarlyExit,swift}.

\paragraph{Token-wise layer skipping.}
Token-wise methods use token representations to decide whether each token executes or skips a layer or sublayer. In SkipGPT, \citet{SkipGPT} place a router before each attention and FFN sublayer. Each router reads the current token state and decides whether to execute or skip its sublayer, allowing different tokens to take different actions at the same sublayer. During router training, they penalize deviation of the average execution rate over tokens and modules from a fixed target. In Mixture-of-Depths, \citet{MixtureOfDepth} set a token capacity at each layer. In D-LLM, \citet{d-llm} instead draw each token's executed depth toward a target, with a separate checkpoint trained for each target budget. \citet{FFNSkipLLM} and \citet{diffskip} skip FFN computation while retaining attention in FFN-SkipLLM and DiffSkip, respectively. In FlexiDepth, \citet{FlexiDepthAdaptive} train layer routers with a penalty on layer execution and restrict skipping to the latter half of the Transformer blocks; the penalty coefficient controls the tendency to skip rather than specifying an execution target for each request. In GateSkip, \citet{WLW} gate attention and FFN sublayers using batch-wide thresholds. In BUDDY, \citet{buddy} change which layers execute across tokens while keeping their number fixed by the request's budget. \citet{RouterTuning} study router-only adaptation in Router-Tuning, and \citet{duo-llm} route tokens among differently sized FFN modules in Duo-LLM. D-LLM's Decision Module and SkipGPT's routers take the token state as input. \method{} also supplies the request's budget at each routed layer.

\section{Method: \method{}}
\label{sec:method}
\subsection{Per-request depth targets}
\label{sec:depth-targets}
The prompt and the first two decode positions of each request execute all layers. Routing starts at the third decode position. Let $\mathcal T_b$ contain the routed positions of request $b$, and let $T_b=|\mathcal T_b|$. At position $t$, $z_{b,t}^{\ell}=1$ if the token executes layer $\ell$ and $0$ if it skips it. We count all $L$ layers, including the first two layers, which always execute. The fraction of layers executed at $t$ and its average over request $b$ are
\begin{equation}
a_{b,t}=\frac{1}{L}\sum_{\ell=1}^{L}z_{b,t}^{\ell},
\qquad
\bar a_b=\frac{1}{T_b}\sum_{t\in\mathcal T_b}a_{b,t},
\qquad (T_b>0).
\label{eq:execution-fraction}
\end{equation}
The budget $\omegaT_b$ sets a target for $\bar a_b$. We compute the absolute budget error for each request with at least one routed position, then average these errors. With $\mathcal B_+=\{b:T_b>0\}$,
\begin{equation}
 \mathcal L_{\mathrm{budget}}=
 \frac{1}{|\mathcal B_+|}\sum_{b\in\mathcal B_+}|\bar a_b-\omegaT_b|.
 \label{eq:active-loss}
\end{equation}
In a 32-layer model, two routed tokens can execute 12 and 20 layers and meet a 50\% request budget exactly. The loss acts on their average. Neither token incurs a separate penalty for its distance from 50\%. The language-model loss guides which layers each token executes. Each request incurs its own absolute error, so another request cannot offset its excess. D-LLM~\citep{d-llm} penalizes each token's deviation from its target depth. SkipGPT~\citep{SkipGPT} averages the fraction of skipped modules across tokens in the training micro-batch and penalizes its deviation from the sparsity target.

\subsection{Routing and executing a token}
\label{sec:routing-token}
\begin{figure}[htbp]
\centering
\includegraphics[width=0.97\textwidth]{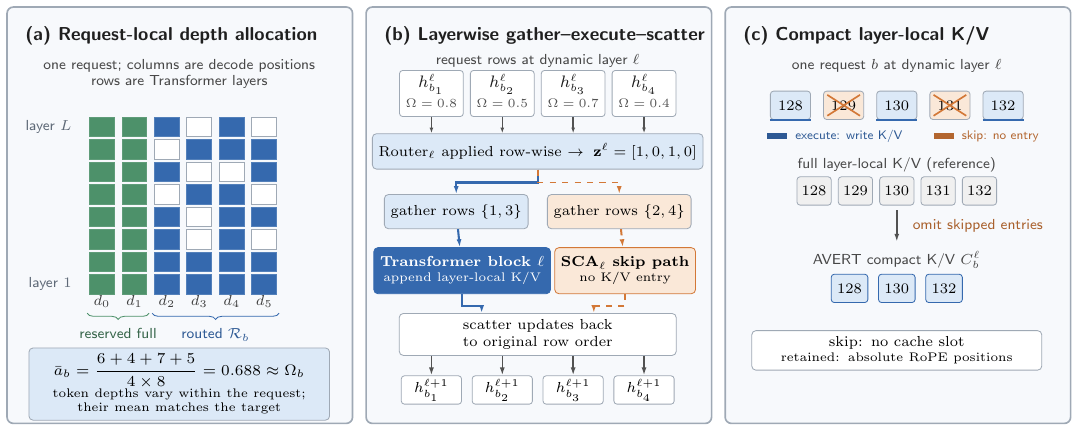}
\vspace{-0.5em}
\caption{\textbf{\method{}.} (a) Tokens in one request can execute different numbers of layers while their average tracks $\omegaT_b$. (b) Tokens selected for block execution are processed together, while skipped tokens pass through SCA. (c) Each layer appends K/V entries only for tokens that execute it.}
\label{fig:avert-arch}
\end{figure}

At each routed layer $\ell$, the router takes the token's current hidden state $h_{b,t}^{\ell}$ and the budget $\omegaT_b$ of its request as input (Figure~\ref{fig:avert-arch}). We call this routing \emph{request-local} because each decision is made independently of other requests in the batch. The router needs no running count of prior executions at inference.
\begin{equation}
 \mathbf{u}_{b,t}^{\ell}=W_{h}^{\ell}h_{b,t}^{\ell}+\mathbf{b}_{h}^{\ell}+W_{\Omega}^{\ell}\omegaT_b,\qquad
 \mathbf{s}_{b,t}^{\ell}=W_{o}^{\ell}\,\phi\!\left(\mathrm{RMSNorm}(\mathbf{u}_{b,t}^{\ell})\right)+\mathbf{b}_{o}^{\ell}\in\mathbb{R}^{2},
 \label{eq:router}
\end{equation}
The hidden-state projection $W_h^\ell$ maps the token state to the router width (4096 to 512 on LLaMA-3.1-8B). The projection $W_\Omega^\ell$ maps the scalar budget to the same width. After RMS normalization and a SiLU activation $\phi$, $W_o^\ell$ produces two scores, one for skipping and one for execution. At inference the higher score determines the action, with ties resolved in favor of execution.

If the token executes the block, it runs both attention and the feed-forward network (FFN) and writes a K/V entry at this layer. If it skips, it bypasses the whole block and passes through a residual Skip Compensation Adapter (SCA).
\begin{equation}
 h_{b,t}^{\ell+1}=
 \begin{cases}
   \mathrm{Block}_{\ell}(h_{b,t}^{\ell}), & z_{b,t}^{\ell}=1,\\
   h_{b,t}^{\ell}+\mathrm{SCA}_{\ell}(h_{b,t}^{\ell}), & z_{b,t}^{\ell}=0.
 \end{cases}
 \label{eq:skip-path}
\end{equation}
SCA is a width-256 bottleneck MLP with a zero-initialized up-projection~\citep{FlexiDepthAdaptive}. The skip path starts as the identity and learns a correction to the token's hidden state. Each router reads the token state produced by the preceding block or skip path, so earlier decisions can influence routing decisions at later layers. We refer to this layer-by-layer decision process as \emph{progressive routing}.

The skip path writes no K/V entry. Future tokens therefore attend only to the shorter compact K/V history at that layer. Omitting the entry saves the current write and reduces stored K/V and the work of later attention. Omitting a K/V entry can also affect predictions for later tokens (Appendix~\ref{app:s1s2-full}). During training, the cross-entropy loss on later tokens can influence routing decisions for earlier tokens. Algorithm~\ref{alg:decode} in Appendix~\ref{app:decode} gives the decode step.

\subsection{Training and batched decoding}
\label{sec:training-batching}
\paragraph{Training with sampled budgets.}
We train on 51,564 Alpaca examples~\citep{alpaca}. In 90\% of micro-batches, each request independently draws a budget uniformly from $[0.4,1.0]$. For the other 10\%, we choose one endpoint at random and assign it to the whole micro-batch. Each request's sampled budget goes to its routers and serves as the target in its budget loss. During training, a straight-through Gumbel--Softmax estimator with temperature 1 yields a hard routing decision in the forward pass and soft gradients in the backward pass~\citep{gumbelSoftmax}. We compute cross-entropy (CE) on supervised answer tokens and optimize the combined objective
\begin{equation}
\mathcal L = \mathcal L_{\mathrm{CE}}
+ \lambda_{\mathrm{budget}}\mathcal L_{\mathrm{budget}}.
\label{eq:full-loss}
\end{equation}
We freeze the backbone and jointly train the routers, SCA, and LoRA adapters~\citep{lora}. LoRA updates the attention query, key, and value projections and the FFN up, down, and gate projections. Under the default configuration, the budget-loss weight $\lambda_{\mathrm{budget}}$ increases from 0 to 10 over the first epoch and stays at 10 thereafter. Appendix~\ref{app:reproducibility} gives the optimization settings, and Appendix~\ref{app:overhead} reports the router and SCA parameter counts.

\paragraph{Decoding requests with different budgets in one batch.}
Training over a range of budgets, together with routing conditioned on each request's own budget, allows a single checkpoint to decode requests with different budgets in the same batch. At each routed layer, \method{} gathers the tokens routed to execution and runs the block on them, while applying SCA to the tokens that skip. It restores outputs from both paths to the original request order before moving to the next layer.

\section{Experiments}
\label{sec:experiments}
\subsection{Setup}
\label{sec:setup}
\paragraph{Models and baselines.}
We evaluate LLaMA-2-7B (multi-head attention, MHA; 32 layers) from \citet{llama2}, LLaMA-3.1-8B (grouped-query attention, GQA; 32 layers), and LLaMA-3.2-1B (GQA, 16 layers) from \citet{LlamaHerd}. We compare with the Alpaca-finetuned Vanilla reference, D-LLM, GateSkip-FFT, GateSkip-LoRA, BUDDY (reimpl.), and AdaSkip. Vanilla and GateSkip-LoRA use our LoRA setting; GateSkip-FFT jointly fine-tunes all backbone and gate parameters, as in \citet{WLW}. D-LLM requires one checkpoint per target, requiring 13 training runs for our sweep. For D-LLM efficiency measurements, we omit K/V entries that its released code computes and masks, keeping its weights and layer decisions unchanged. AdaSkip comparisons disable the online window during decoding. Baseline implementations are described in Appendix~\ref{app:reproducibility}; Table~\ref{tab:budget-interface} compares their budget controls. A reduced Qwen2.5-3B evaluation is in Appendix~\ref{app:qwen25}.

\begin{table}[htbp]
\centering
\caption{How the methods use compute budgets.}
\label{tab:budget-interface}
\small
\setlength{\tabcolsep}{4pt}
\renewcommand{\arraystretch}{1.17}
\renewcommand{\tabularxcolumn}[1]{m{#1}}
\begin{tabularx}{\linewidth}{@{}l >{\raggedright\arraybackslash}m{.14\linewidth} >{\raggedright\arraybackslash}X >{\centering\arraybackslash}m{.235\linewidth}@{}}
\toprule
\multicolumn{1}{@{}c}{\makecell[c]{Method}} &
\multicolumn{1}{c}{\makecell[c]{Budget set for}} &
\multicolumn{1}{c}{\makecell[c]{What the budget controls}} &
\multicolumn{1}{c@{}}{\makecell[c]{Mixed-budget inference\\batches explicitly supported?}} \\
\midrule
AdaSkip & Inference run & The budget determines a fixed subset of attention and FFN sublayers shared by all tokens. & No \\
GateSkip & Batch & The budget selects a quantile of the batch-wide live tokens' scores to determine which tokens execute each sublayer. & No \\
D-LLM & Training run & Each token is penalized for deviating from the training target. & No \\
MoD & Training run & The budget sets each routed layer's capacity, limiting how many tokens execute it during training. & No \\
SkipGPT & Training run & The training target sets the fraction of attention and FFN modules to execute on average across tokens in a micro-batch. & No \\
FlexiDepth & Training run & A penalty coefficient controls the tendency to skip. Users cannot specify a target execution fraction. & No \\
L2A & Request & The budget sets a target for the combined cost of layers and attention heads and guides when the model switches from reasoning to answering. & No \\
BUDDY & Request & The budget fixes how many layers each token executes. & No \\
\textbf{\method{}} & \textbf{Request} & The budget sets a target for the average fraction of layers executed across a request's routed tokens. & \textbf{Yes} \\
\bottomrule
\end{tabularx}
\end{table}

\paragraph{Tasks and metrics.}
We evaluate 1,000 held-out Alpaca examples. Open-ended Alpaca instructions can have multiple valid answers that differ from the reference. We therefore use Prometheus-2 (1--5)~\citep{prometheus} as the primary quality metric to assess responses against the instruction and a fixed scoring rubric, alongside ROUGE-L~\citep{rouge} and BLEU-4~\citep{bleu} for lexical overlap. GSM8K uses 8-shot prompting and flexible answer-extraction accuracy on 1,319 questions~\citep{TrainingWordProblems}. Zero-shot IFEval reports strict prompt and instruction accuracies on 541 prompts with 834 instructions, using rule-based checks~\citep{ifeval}. These six metrics and all system measurements use full-compute prefill and skip only during decode. Alpaca evaluation examples are excluded from training; GSM8K and IFEval are not used for fine-tuning. The auxiliary zero-shot PIQA probe uses a separately trained prefill-skipping checkpoint (Appendix~\ref{app:reproducibility}).

\paragraph{Budgets and measurements.}
We evaluate 13 targets from 0.40 to 1.00 in increments of 0.05. Measured execution fraction counts whole blocks for \method{}, D-LLM, and BUDDY, and attention/FFN sublayers for GateSkip and AdaSkip. For \method{}, all $L$ layers are counted at each routed position (Section~\ref{sec:depth-targets}). We compute $e_b=|\bar a_b-\omegaT_b|$ separately for each request. For each backbone and budget, we report the mean of $e_b$ over requests assigned that budget, in percentage points.

System measurements use one NVIDIA A800 80GB GPU, bfloat16 inference, a batch size (BS) of 16, 128 full-depth prefill tokens, and 128 output tokens per request unless specified otherwise. Synchronized decode TPOT includes the two full-depth decode positions, routing, gather/scatter, SCA, and block execution. We also report peak decode GPU memory (VRAM). Workloads use fixed requests and synchronized PyTorch decoding; timing and memory protocols are in Appendix~\ref{app:reproducibility}. Mixed-versus-split comparisons use the same checkpoint, request--budget pairs, and decoding implementation (Appendix~\ref{app:mixed-split}).

\subsection{Quality and efficiency}
\label{sec:exp-main}
\begin{table}[t]
\centering
\caption{Main results at the near-full target ($\omegaT=1.0$) and the aggressive target ($\omegaT=0.4$). IFEval-SP/SI are strict prompt/instruction accuracies; GSM8K-Flex is flexible answer-extraction accuracy. Best dynamic result per setting is in \textbf{bold}, using unrounded values.}
\label{tab:main}
\scriptsize
\setlength{\tabcolsep}{2.8pt}
\resizebox{\textwidth}{!}{%
\begin{tabular}{ll cc cc cc cc cc cc}
\toprule
& & \multicolumn{2}{c}{Prom.} & \multicolumn{2}{c}{ROUGE-L}
  & \multicolumn{2}{c}{BLEU-4} & \multicolumn{2}{c}{GSM8K-Flex}
  & \multicolumn{2}{c}{IFEval-SP} & \multicolumn{2}{c}{IFEval-SI} \\
\cmidrule(lr){3-4}\cmidrule(lr){5-6}\cmidrule(lr){7-8}\cmidrule(lr){9-10}
\cmidrule(lr){11-12}\cmidrule(lr){13-14}
Model & Method & 1.0 & 0.4 & 1.0 & 0.4 & 1.0 & 0.4 & 1.0 & 0.4 & 1.0 & 0.4 & 1.0 & 0.4 \\
\midrule
\multirow{7}{*}{\makecell[l]{LLaMA-2-7B\\(MHA, 32L)}}
  & Vanilla & 3.39 & -- & 0.278 & -- & 0.116 & -- & 0.116 & -- & 0.246 & -- & 0.354 & -- \\
  & GateSkip-FFT & 3.08 & 1.10 & 0.261 & 0.159 & 0.104 & 0.022 & 0.024 & 0.001 & 0.142 & 0.059 & 0.252 & 0.182 \\
  & GateSkip-LoRA & 3.32 & 1.00 & 0.255 & 0.000 & 0.109 & 0.000 & 0.033 & 0.002 & 0.202 & 0.046 & 0.301 & 0.133 \\
  & D-LLM & \textbf{3.40} & 1.51 & 0.281 & 0.190 & 0.125 & 0.036 & 0.120 & 0.011 & 0.255 & 0.081 & 0.366 & 0.177 \\
  & AdaSkip & 3.38 & 1.01 & 0.277 & 0.034 & 0.116 & 0.002 & 0.116 & 0.003 & 0.244 & 0.061 & 0.352 & 0.131 \\
  & BUDDY (reimpl.) & 3.12 & 1.39 & \textbf{0.291} & 0.170 & 0.122 & 0.034 & 0.064 & 0.015 & 0.196 & 0.068 & 0.302 & 0.157 \\
  & \textbf{\method{} (Ours)} & \textbf{3.40} & \textbf{2.38} & 0.279 & \textbf{0.221} & \textbf{0.126} & \textbf{0.080} & \textbf{0.121} & \textbf{0.017} & \textbf{0.275} & \textbf{0.159} & \textbf{0.398} & \textbf{0.255} \\
\midrule
\multirow{7}{*}{\makecell[l]{LLaMA-3.1-8B\\(GQA, 32L)}}
  & Vanilla & 3.66 & -- & 0.280 & -- & 0.123 & -- & 0.428 & -- & 0.325 & -- & 0.436 & -- \\
  & GateSkip-FFT & 3.00 & 1.02 & 0.251 & 0.113 & 0.096 & 0.019 & 0.083 & 0.006 & 0.196 & 0.056 & 0.330 & 0.160 \\
  & GateSkip-LoRA & 3.58 & 1.05 & 0.281 & 0.140 & 0.124 & 0.014 & 0.317 & 0.005 & 0.257 & 0.048 & 0.384 & 0.132 \\
  & D-LLM & \textbf{3.66} & 1.62 & 0.284 & 0.173 & 0.124 & 0.050 & 0.422 & 0.005 & 0.359 & 0.052 & 0.460 & 0.146 \\
  & AdaSkip & \textbf{3.66} & 1.01 & 0.281 & 0.009 & 0.124 & 0.001 & \textbf{0.428} & 0.000 & 0.324 & 0.061 & 0.438 & 0.091 \\
  & BUDDY (reimpl.) & 3.29 & 1.48 & \textbf{0.298} & 0.170 & \textbf{0.126} & 0.035 & 0.176 & 0.016 & 0.242 & 0.070 & 0.354 & 0.163 \\
  & \textbf{\method{} (Ours)} & 3.65 & \textbf{2.02} & 0.281 & \textbf{0.199} & 0.124 & \textbf{0.061} & 0.425 & \textbf{0.020} & \textbf{0.362} & \textbf{0.092} & \textbf{0.471} & \textbf{0.189} \\
\midrule
\multirow{7}{*}{\makecell[l]{LLaMA-3.2-1B\\(GQA, 16L)}}
  & Vanilla & 2.93 & -- & 0.250 & -- & 0.098 & -- & 0.044 & -- & 0.128 & -- & 0.224 & -- \\
  & GateSkip-FFT & 2.66 & 1.06 & 0.240 & 0.065 & 0.085 & 0.006 & 0.023 & 0.001 & \textbf{0.142} & 0.070 & \textbf{0.249} & 0.145 \\
  & GateSkip-LoRA & 1.95 & 1.08 & 0.214 & 0.061 & 0.069 & 0.005 & 0.026 & 0.003 & 0.103 & 0.052 & 0.199 & 0.144 \\
  & D-LLM & 2.93 & 1.26 & 0.247 & 0.166 & 0.098 & 0.045 & \textbf{0.045} & 0.011 & 0.120 & 0.074 & 0.222 & 0.152 \\
  & AdaSkip & 2.92 & 1.00 & 0.249 & 0.001 & 0.097 & 0.000 & 0.044 & 0.000 & 0.126 & 0.061 & 0.225 & 0.134 \\
  & BUDDY (reimpl.) & 2.33 & 1.17 & \textbf{0.250} & 0.133 & 0.091 & 0.018 & 0.018 & \textbf{0.021} & 0.085 & 0.081 & 0.186 & \textbf{0.165} \\
  & \textbf{\method{} (Ours)} & \textbf{2.94} & \textbf{1.44} & 0.250 & \textbf{0.187} & \textbf{0.098} & \textbf{0.057} & 0.045 & 0.017 & 0.133 & \textbf{0.085} & 0.239 & 0.158 \\
\bottomrule
\end{tabular}%
}
\end{table}

\begin{figure}[t]
\centering
\includegraphics[width=\textwidth]{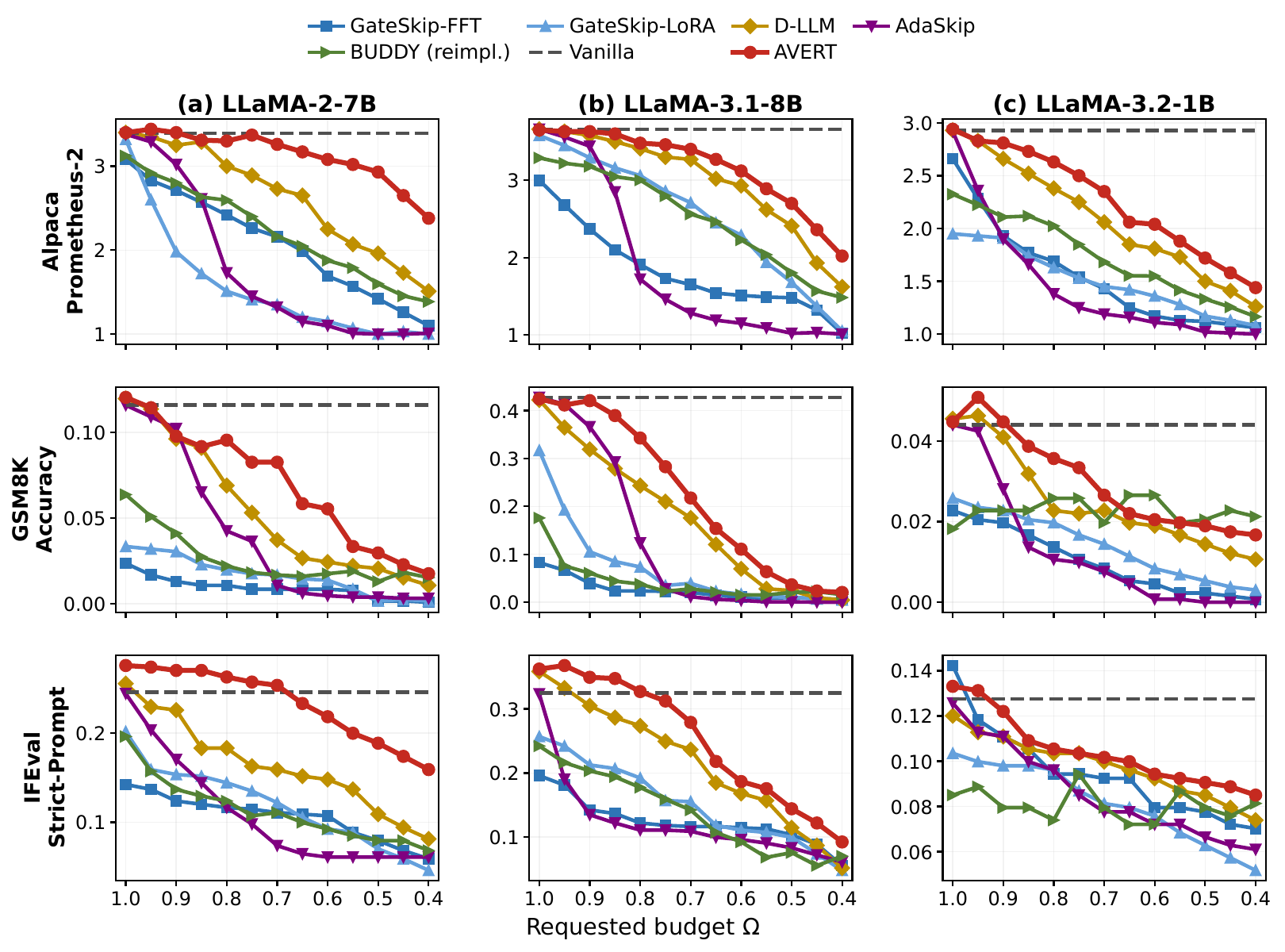}
\vspace{-0.6em}
\caption{\textbf{Quality across the full 13-point budget sweep.} Columns are backbones; rows report Prometheus-2 on Alpaca, GSM8K flexible accuracy, and IFEval strict-prompt accuracy. Each curve connects the 13 requested targets, and dashed lines mark Vanilla. Remaining metrics are in Appendix~\ref{app:main-full}.}
\label{fig:quality-grid}
\end{figure}

Table~\ref{tab:main} reports the two endpoint targets, and Figure~\ref{fig:quality-grid} shows the full sweep. Across three backbones, six main metrics, and 13 targets, \method{} is best or tied for best in 206/234 comparisons. At $\omegaT=1.0$, it remains close to Vanilla and the strongest baselines. \method{} still uses normal routing, and some tokens may skip layers. At $\omegaT=0.4$, \method{} achieves the best or tied-best result in 16 of the 18 comparisons across three backbones and six quality metrics. LLaMA-3.1-8B can degrade more steeply than LLaMA-2-7B at low budgets despite higher full-compute quality. Complete results, including the auxiliary PIQA probe, are in Appendix~\ref{app:main-full}.

At full budget on LLaMA-3.1-8B, BUDDY scores 0.2980/0.1264 on ROUGE-L/BLEU-4 versus 0.2808/0.1238 for \method{}, but Prometheus-2 scores are 3.29 versus 3.65. Asked for two synonyms, BUDDY repeats \emph{enraged} and still receives ROUGE-L 0.900. Appendix~\ref{app:flexidepth-answers} gives a similar FlexiDepth example. Table~\ref{tab:additional-baselines} in Appendix~\ref{app:skipgpt-budget} compares SkipGPT and FlexiDepth at their measured execution fractions. Same-checkpoint forced-full evaluation and five training seeds are reported in Appendices~\ref{app:force-full} and~\ref{app:seeds}.

\begin{figure}[t]
\centering
\includegraphics[width=\textwidth]{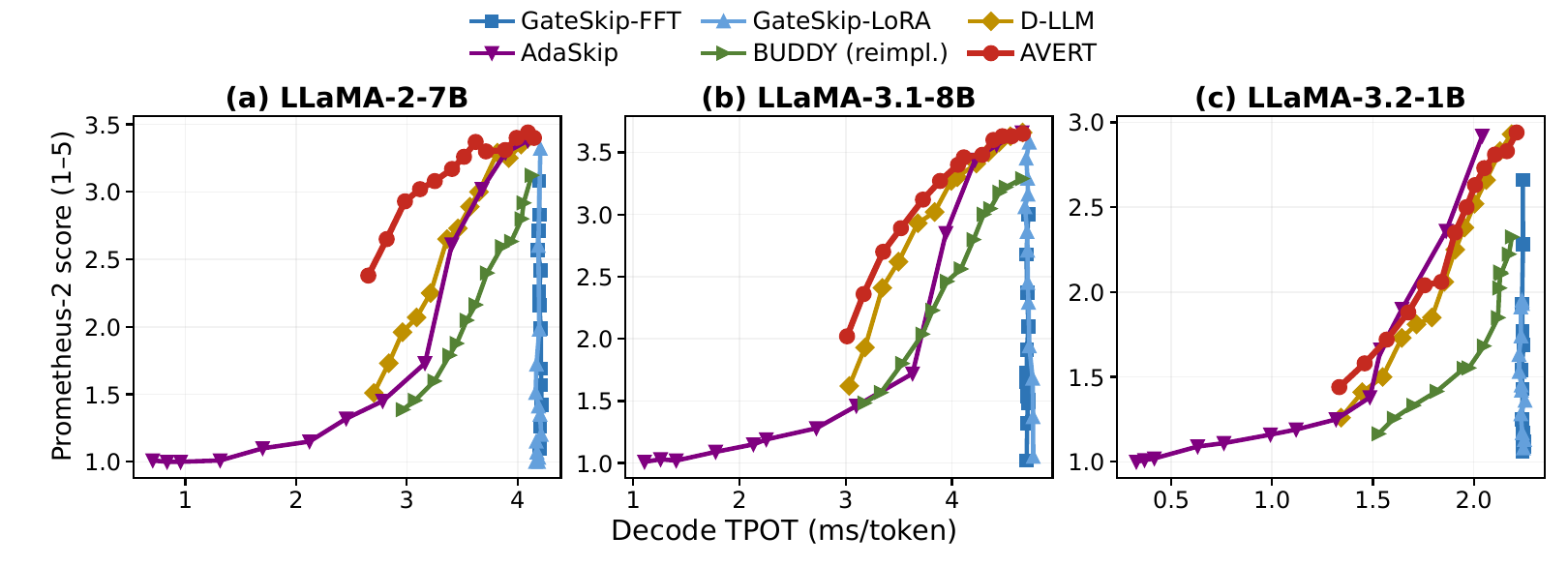}
\caption{\textbf{Prometheus-2 versus measured decode TPOT.}
Each point shows a method's measured TPOT on the horizontal axis and Prometheus-2 score on the vertical axis at one requested budget.
Each curve connects the 13 evaluated budgets for one method on a fixed backbone.}
\label{fig:prometheus-tpot-tradeoff}
\end{figure}

\paragraph{Quality versus measured decode time.}
Figure~\ref{fig:prometheus-tpot-tradeoff} plots Prometheus-2 against decode TPOT. Quality and timing use their respective evaluation protocols, with points paired by method, backbone, and requested target. 

Against AdaSkip, we select three pairs per backbone from the existing quality sweeps before timing, with Prometheus-2 gaps of at most 0.06. \method{} has lower TPOT in seven of nine pairs. For example, on LLaMA-3.1-8B, \method{} at $\omegaT=0.55$ scores 2.89 at 3.519 ms/token; AdaSkip at $\omegaT=0.85$ scores 2.85 at 3.942 ms/token. Appendix~\ref{app:adaskip-efficiency} gives all pairs and memory results; Appendix~\ref{app:existing-quality-cost} compares the existing operating points across methods. Comparisons by measured execution fraction and sensitivity to prompt length are in Appendices~\ref{app:realized-activation-efficiency} and~\ref{app:prompt-length}. Detailed router, dispatch, SCA, and attention/K/V timings are in Appendix~\ref{app:components}.

\subsection{Per-request budget control and mixed batching}
\label{sec:exp-batching}
At a 60\% budget, one LLaMA-3.1-8B response uses between 2 and 31 layers across its 125 routed tokens. It executes 2,453 blocks, averaging 19.624 layers per token and 61.325\% execution. Token depth varies within all 480 responses in this experiment; 460 responses (95.83\%) have mean execution within two percentage points of their targets. The maximum absolute request error does not exceed 7.5 points (Appendix~\ref{app:request-depth}). At a fixed budget, tokens in higher quartiles of next-token prediction entropy execute more layers on average (Appendix~\ref{app:routing-behavior}).

At BS=32, we average the absolute budget errors of requests assigned the same target, separately for each backbone. Across all tested targets and three backbones, the largest of these averages is 0.97 percentage points for single-budget batches and 1.15 points for mixed-budget batches (Appendix~\ref{app:mixed-split}). Across five additional targets between the main evaluation points, the largest average is 0.90 points (Appendix~\ref{app:offgrid}).

In a 32-layer BUDDY model, targets 0.940, 0.945, and 0.950 all select 30 layers per token, so the count changes in steps of $1/32$. In \method{}, one additional block execution across $T_b$ routed tokens changes the request average by $1/(32T_b)$. Users can supply any budget in the trained interval.

GateSkip-FFT derives its sublayer thresholds from the batch-wide live-token distribution. On LLaMA-3.1-8B at $\omegaT=0.4$, its execution fraction is 1.000 at BS=1, 0.632 at BS=2, approximately 0.298 at BS=8/16, and 0.289 at BS=32. Prometheus-2 ranges from 3.01 to 1.02. Quantile thresholding does not explicitly retain an exact integer $k$, so its batch-wide execution fraction can differ from the target. \method{} remains stable across its own batch-size sweep (Appendix~\ref{app:bs-full}).

We compare mixed and budget-separated batches using the same 128 requests, with budgets 1.0, 0.8, 0.6, and 0.4, and average ten runs. With equal request counts at each budget, both schedules form four full batches of 32: TPOT differs by at most 0.15\% and total time by at most 0.24\%. Under moderate skew, where the 128 requests are distributed unevenly across the four budgets, grouping by budget requires five batches with 80\% average fill; under strong skew, it requires six batches with 66.7\% fill. Mixing budgets still forms four full batches.

Across the three backbones and skewed workloads, mixing budgets reduces TPOT by 4.3--21.2\% and total wall-clock time by 4.1--20.9\%. For LLaMA-3.1-8B with mostly low-budget requests, total time including prefill falls from 60.60 to 52.15 seconds. Appendix~\ref{app:mixed-split} gives workload composition, batch packing, and memory results.

\subsection{Ablations}
\label{sec:exp-ablation}
\paragraph{Budget input.}
We remove $W_\Omega^\ell$ from every router while keeping budget sampling and the request-average loss unchanged. Without the budget input, users can no longer set a budget for any request at inference. The target varies during training, but the routers no longer receive it. On LLaMA-3.1-8B Alpaca, the mean execution fraction is 74.59\%, and the mean Prometheus-2, ROUGE-L, and BLEU-4 scores are 3.4470, 0.2650, and 0.1115, respectively. Standard \method{} at a 75\% budget scores 3.46, 0.2661, and 0.1122. Keeping the budget input but removing its penalty ($\lambda_{\mathrm{budget}}=0$) gives 99.60--100\% execution at the four tested targets (Appendix~\ref{app:penalty}).

\begin{figure}[t]
\centering
\includegraphics[width=\textwidth]{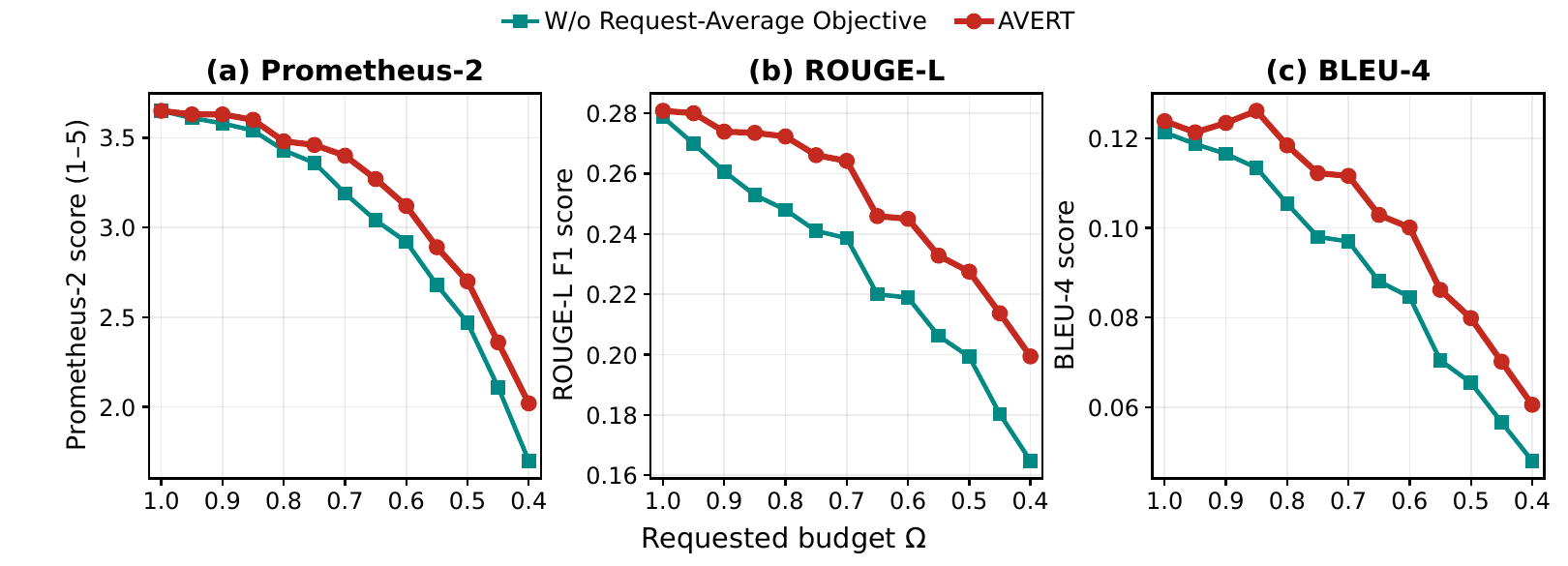}
\caption{\textbf{Request-average objective ablation.} On LLaMA-3.1-8B, ``W/o Request-Average Objective'' uses the per-token loss.}
\label{fig:app-request-average-objective}
\end{figure}

\paragraph{Request-average objective.}
Figure~\ref{fig:app-request-average-objective} compares request-average and per-token objectives with the same architecture and decoding implementation. The per-token objective penalizes each token's depth deviation separately. Across 13 targets, mean execution differs by at most 0.71 percentage points. The request-average objective gives higher Prometheus-2 at 12 targets and ties at one; ROUGE-L and BLEU-4 are higher at all 13 targets.

In the fifth epoch, the mean budget loss used for parameter updates was 0.0070 with the request-average objective and 0.0164 with the per-token objective. Cross-entropy is lower with the request-average objective in all five epochs (Appendix~\ref{app:training-curves}).

\paragraph{Layer-by-layer decisions.}
The one-shot ablation computes all dynamic-layer decisions from the same hidden state before the first dynamic layer. Across 13 targets, its mean execution differs from \method{} by at most 0.92 percentage points. Layer-by-layer routing gives higher Prometheus-2 at 12 targets and higher BLEU-4 at 10 (Appendix~\ref{app:routing-objective-ablations}). With the same reference prefixes across budgets, \method{} sometimes executes a layer at a lower budget but skips it at a higher budget. Its routers receive the budget directly and make decisions layer by layer from the updated token states, allowing the selected layers to change with the budget. BUDDY instead scores layers once per decode step; changing the budget changes $k$ rather than the scores, so its lower-budget selections are subsets of its higher-budget selections for a fixed layer ranking (Appendix~\ref{app:routing-behavior}). Comparing block execution with skipping through SCA, the generated token's rank changes less at positions the router skips than at positions it executes. This indicates that the routers tend to skip layers with a smaller effect on the current prediction (Appendix~\ref{app:logit-lens}).

\paragraph{Skip-path adapter.}
At $\omegaT=0.4$ on LLaMA-3.1-8B, full \method{} and a separately trained no-SCA model execute fractions 0.4067 and 0.4078, respectively. Prometheus-2 is 2.02 with SCA and 1.41 without it; D-LLM scores 1.62. Appendix~\ref{app:sca-full} gives the full sweep.

\paragraph{K/V omission.}
Retaining K/V entries for skipped tokens preserves attention history and can improve quality. Omitting them reduces cache storage and later attention work, lowering peak decode VRAM and generally reducing TPOT at aggressive budgets (Appendix~\ref{app:s1s2-full}).

\section{Limitations and Conclusion}
\label{sec:limitations}
\method{} can lose accuracy at aggressive budgets, particularly on GSM8K. The budget is chosen at the start of a request and remains fixed throughout its response. We have not yet implemented budget changes during generation.

Our main implementation optimizes decoding while keeping prefill at full depth. Evaluation beyond LLaMA is limited to Qwen2.5-3B at three budgets ($0.4$, $0.7$, and $1.0$), with BUDDY and Vanilla as references (Appendix~\ref{app:qwen25}).

\method{} uses a request-average depth objective to match each request's budget while allowing different tokens to execute different numbers of layers. Its request-local progressive routing makes layer decisions from the current token state and that request's budget. One checkpoint supports requests with different budgets in the same batch. Our experiments show improved generation quality at comparable execution fractions, lower TPOT when mixing budgets avoids underfilled batches, and reduced cache storage and peak decode VRAM through K/V omission.

\section*{Reproducibility Statement}
Section~\ref{sec:method} defines the method; Appendix~\ref{app:reproducibility} gives training and evaluation settings. The code provides AVERT training, evaluation, data splits, and our BUDDY reimplementation: \url{https://anonymous.4open.science/r/AVERT-1F8B}. Base checkpoints are obtained separately. Other baselines use public releases with the execution changes described in Appendix~\ref{app:reproducibility}.

\renewcommand{\bibfont}{\small}
\setlength{\bibsep}{3pt plus 1pt}
\bibliographystyle{plainnat}
\bibliography{ref}

\appendix
\raggedbottom
\small

\section{Decoding with layer-by-layer routing and compact K/V caches}
\label{app:decode}
At a single-token decode step, active request $b$ has decode position $r_b$ within the request and absolute sequence position $p_b$. Index $r_b=0$ is the first single-token step after prefill: the token produced by prefill is the input to this step. The reserved positions are $r_b=0$ and $r_b=1$, the first two \emph{decode} inputs rather than prompt positions. Algorithm~\ref{alg:decode} also handles ragged batches, where different active requests may have different $r_b$ values. Each request has its own K/V history $C_b^\ell$ at layer $\ell$. Batched attention temporarily pads the selected requests' histories and masks the padded tail. The cache stores only valid entries. Rotary position embeddings (RoPE) use the original absolute token positions.

\begin{algorithm}[H]
\caption{Request-local progressive decode with compact layer-local K/V}
\label{alg:decode}
\small
\begin{minipage}{0.97\linewidth}
\textbf{Input:} active requests $\mathcal B$; hidden rows $h_b^1$; budgets $\omegaT_b$; decode-input indices $r_b$; absolute positions $p_b$; persistent caches $C_b^\ell$.\par
\vspace{0.25em}
\begin{enumerate}
\setlength{\itemsep}{0.18em}
\setlength{\parskip}{0pt}
\item For layers $\ell=1,\ldots,L$, form the forced-execute set
$F^\ell=\mathcal B$ if $\ell\leq2$, and
$F^\ell=\{b\in\mathcal B:r_b<2\}$ otherwise.
\item For each $b\in\mathcal B\setminus F^\ell$, compute
$z_b^\ell=\mathrm{Router}_\ell(h_b^\ell,\omegaT_b)$ independently. Set
$E^\ell=F^\ell\cup\{b:z_b^\ell=1\}$ and $S^\ell=\mathcal B\setminus E^\ell$.
\item Gather rows in $E^\ell$. Apply RoPE at each absolute position $p_b$, attend to the compact history $C_b^\ell$, execute attention and FFN, and append the new K/V entry to $C_b^\ell$.
\item For every $b\in S^\ell$, set
$h_b^{\ell+1}=h_b^\ell+\mathrm{SCA}_\ell(h_b^\ell)$ and leave $C_b^\ell$ unchanged.
\item Scatter both paths back to the original request order before processing layer $\ell+1$.
\end{enumerate}
\textbf{Output:} apply final normalization and the LM head to $h_b^{L+1}$ for every active request.
\end{minipage}
\end{algorithm}

The objective and routing/depth analyses exclude positions with $r_b<2$; wall-clock decode measurements include them.

\section{Complete quality results across all reported metrics}
\label{app:main-full}
We give the complete 13-point Alpaca/GSM8K sweeps, the auxiliary PIQA prefill-skipping probe, and the full IFEval strict sweeps for each architecture.

At $\omegaT=0.4$, \method{} is best or tied for best in 16 of the 18 main model--metric comparisons. At $\omegaT=0.7$, it is best or tied for best in all 21 comparisons including the auxiliary PIQA probe. Across all 13 targets, it is best or tied for best in 206/234 comparisons on the six main metrics and 240/273 when PIQA is included.

\begin{figure}[!htbp]
\centering
\includegraphics[width=0.96\linewidth]{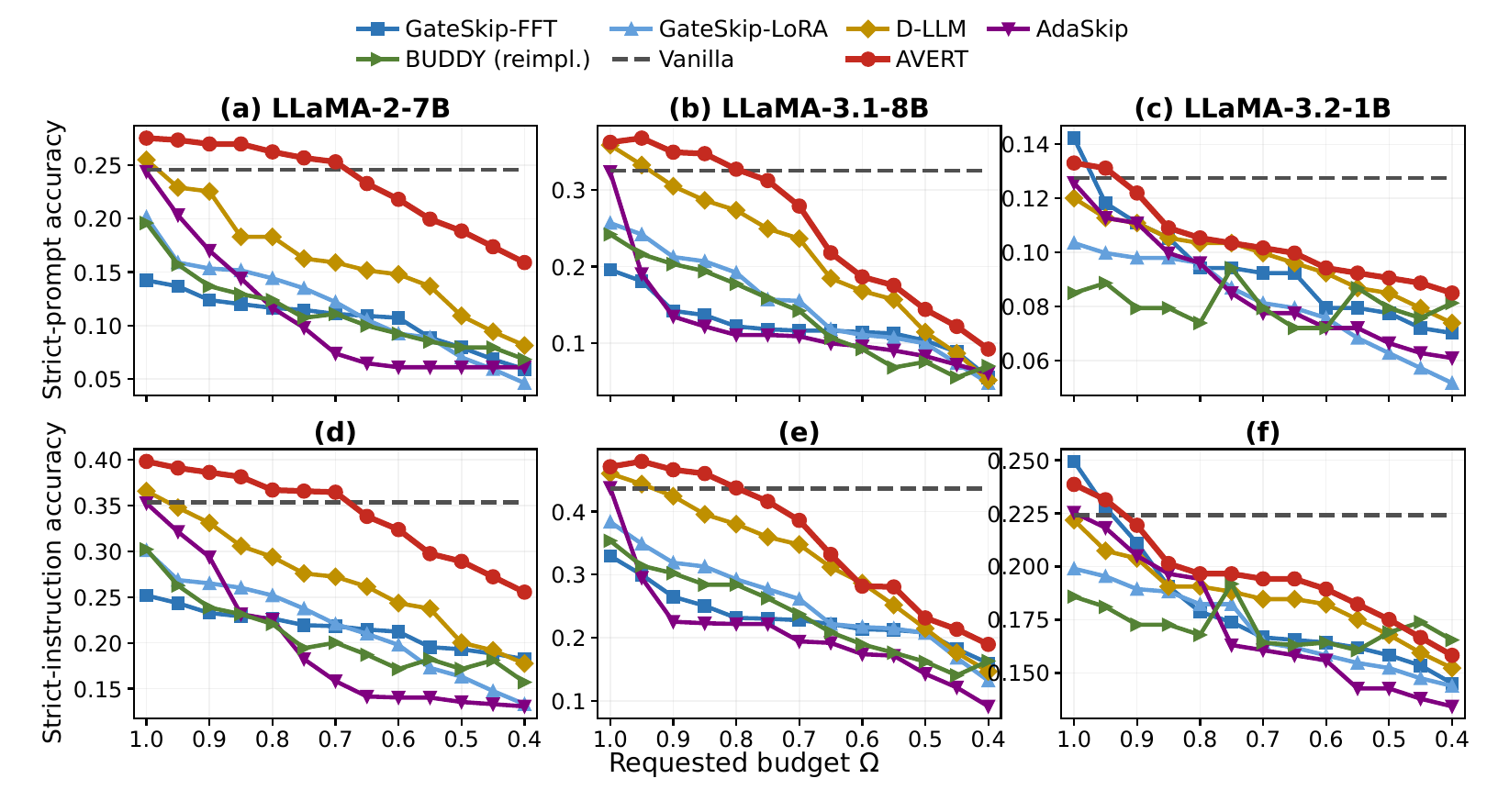}
\caption{Full IFEval strict prompt/instruction sweeps.}
\label{fig:app-ifeval}
\end{figure}

\begin{figure}[!htbp]
\centering
\includegraphics[width=0.96\linewidth]{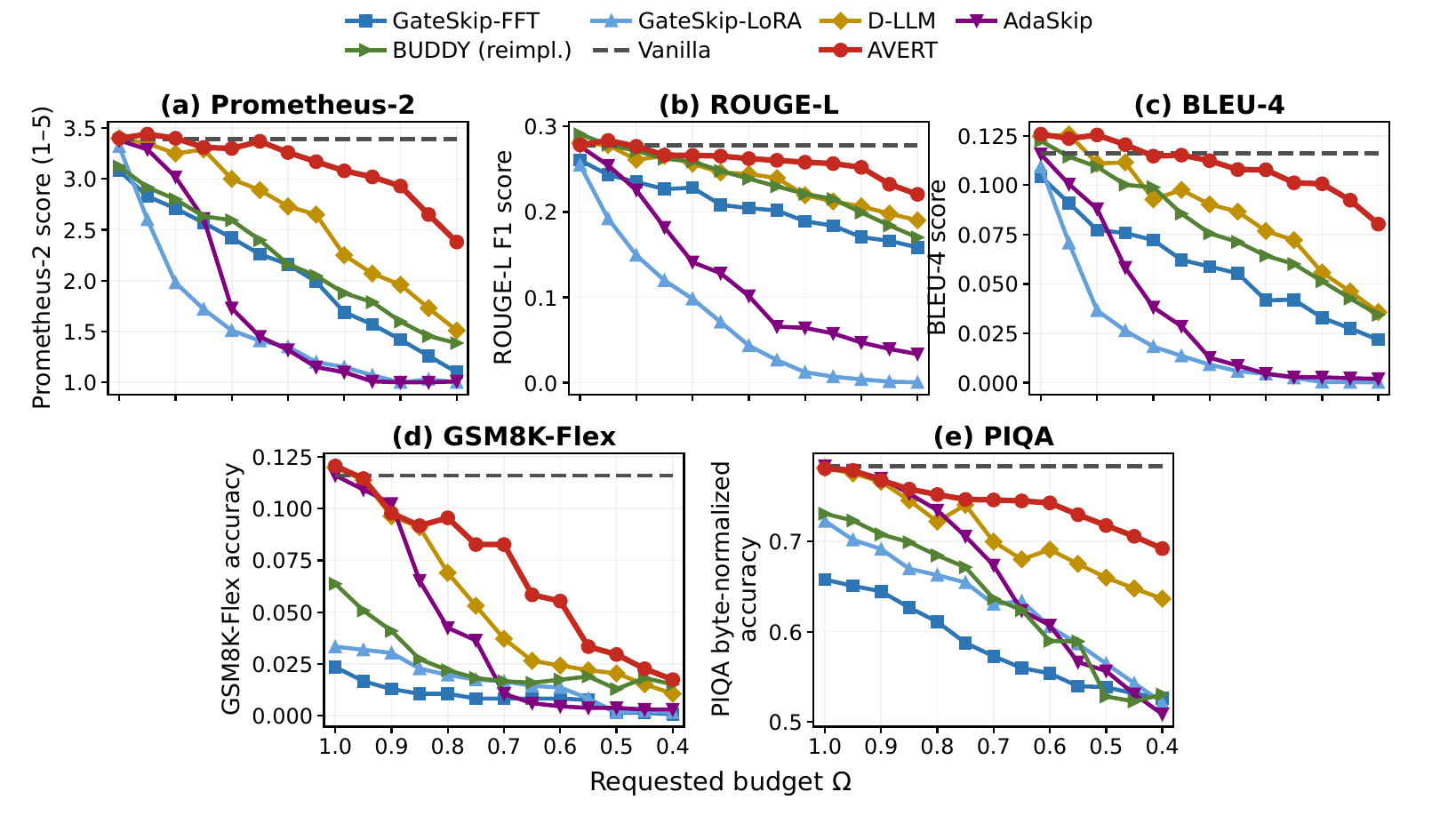}
\caption{\textbf{LLaMA-2-7B (MHA), Alpaca/GSM8K plus auxiliary PIQA metrics.}}
\label{fig:app-llama2-all}
\end{figure}

\begin{figure}[!htbp]
\centering
\includegraphics[width=0.96\linewidth]{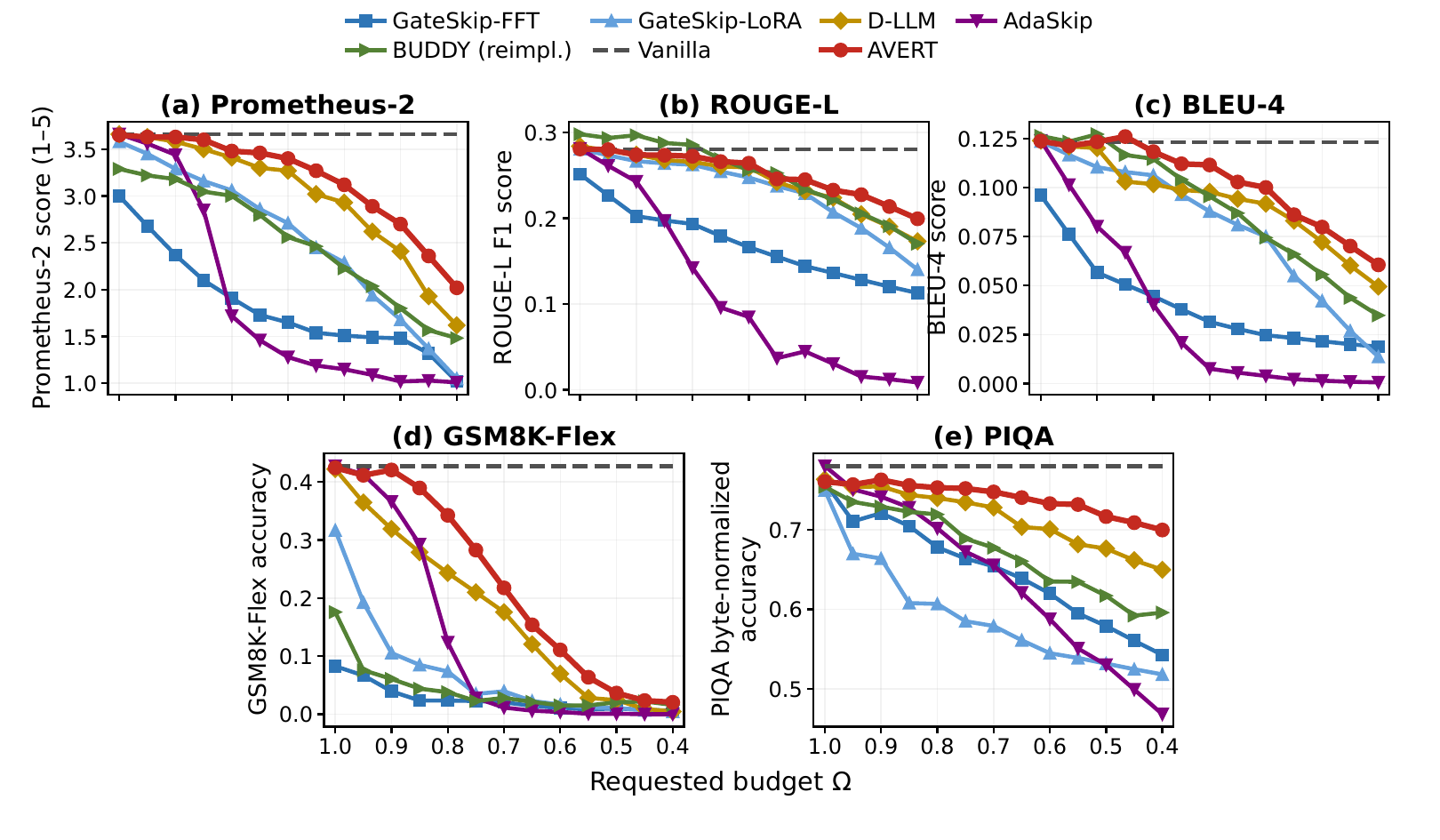}
\caption{\textbf{LLaMA-3.1-8B (GQA), Alpaca/GSM8K plus auxiliary PIQA metrics.}}
\label{fig:app-llama31-all}
\end{figure}

\begin{figure}[!htbp]
\centering
\includegraphics[width=0.96\linewidth]{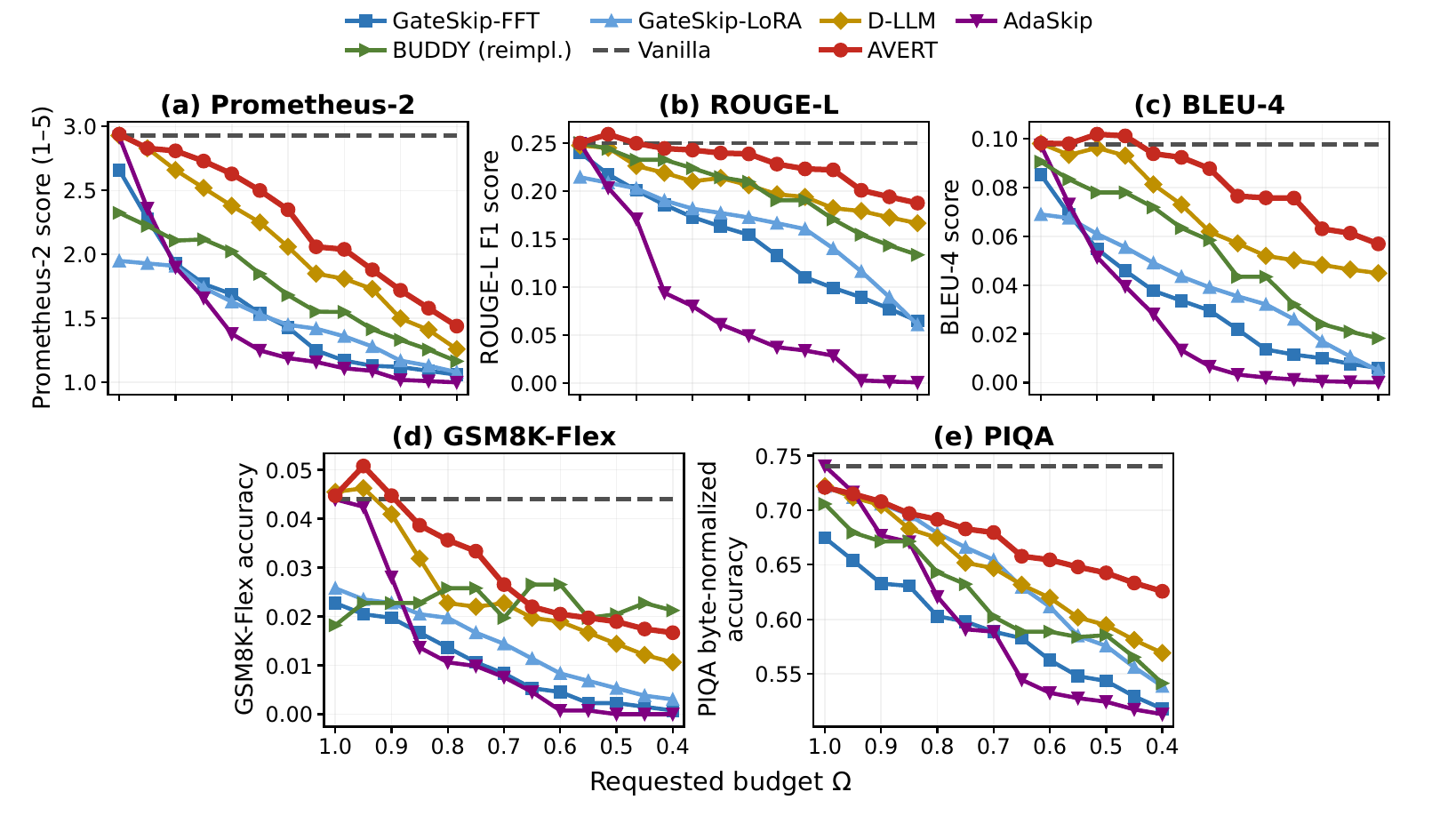}
\caption{\textbf{LLaMA-3.2-1B (GQA, 16L), Alpaca/GSM8K plus auxiliary PIQA metrics.}}
\label{fig:app-llama32-all}
\end{figure}

\subsection{Same-checkpoint forced-full evaluation}
\label{app:force-full}
We retain the trained LLaMA-3.1-8B AVERT weights and evaluate target $1$ with normal routing and with every Transformer block forced to execute (Table~\ref{tab:rebuttal-full}).
Strict-prompt IFEval is $36.23\%$ for both paths, and the largest metric difference is $0.37$ percentage points.
Forced-full AVERT retains a $3.70$-point strict-prompt advantage over the separately adapted Vanilla reference.

\begin{table}[!htbp]
\centering
\caption{\textbf{IFEval full-depth control on LLaMA-3.1-8B.} All scores are percentages; SP/SI and LP/LI denote strict and loose prompt/instruction accuracy.}
\label{tab:rebuttal-full}
\begin{tabular}{lrrrr}
\toprule
Evaluation & SP & SI & LP & LI \\
\midrule
Vanilla & 32.53 & 43.65 & 34.94 & 47.00 \\
AVERT, target $1$ & 36.23 & 47.12 & 38.26 & 50.72 \\
Same AVERT, forced full & 36.23 & 47.00 & 37.89 & 50.60 \\
\bottomrule
\end{tabular}
\end{table}

\FloatBarrier
\section{Full K/V omission-versus-retention ablation across architectures}
\label{app:s1s2-full}

\begin{figure}[!htbp]
\centering
\includegraphics[width=0.91\linewidth]{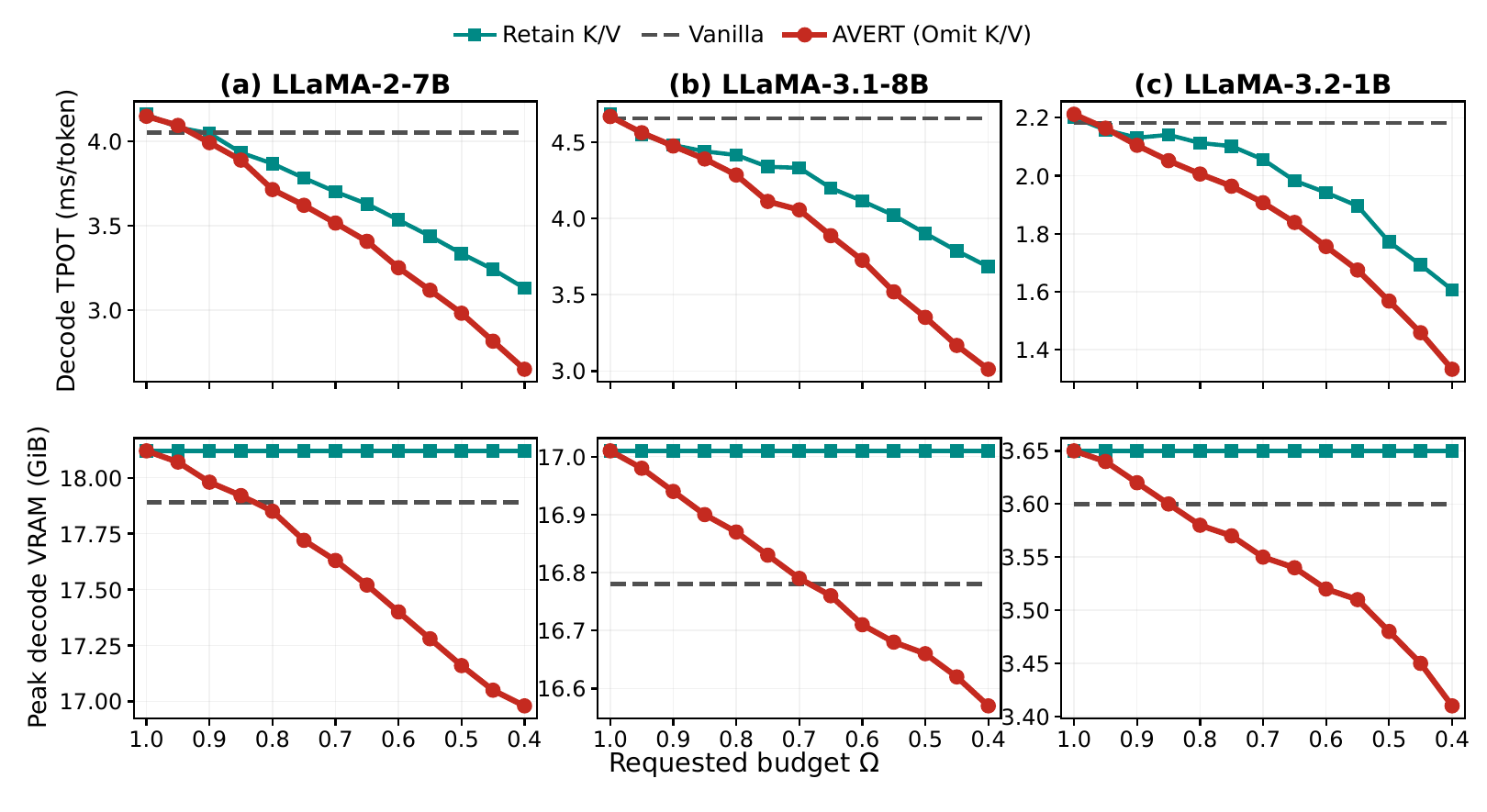}
\caption{\textbf{Decode TPOT and peak VRAM with K/V omission and retention across the three LLaMA backbones.}}
\label{fig:efficiency-grid}
\end{figure}

\begin{figure}[!htbp]
\centering
\includegraphics[width=0.94\linewidth]{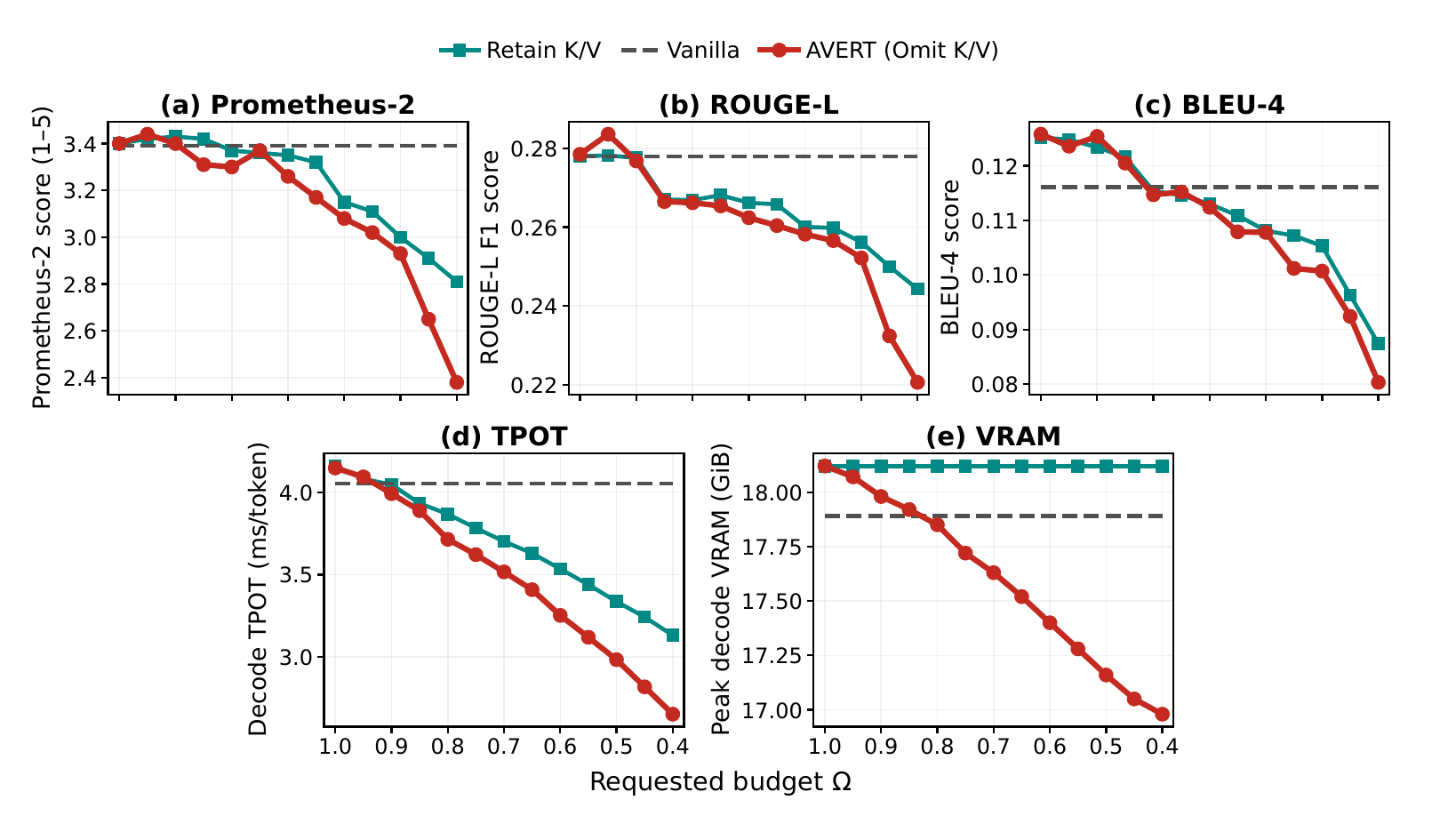}
\caption{\textbf{K/V omission versus retention on LLaMA-2-7B (MHA).} MHA yields the largest cache-memory benefit from omitted K/V entries.}
\label{fig:app-s1s2-llama2}
\end{figure}

\begin{figure}[!htbp]
\centering
\includegraphics[width=0.94\linewidth]{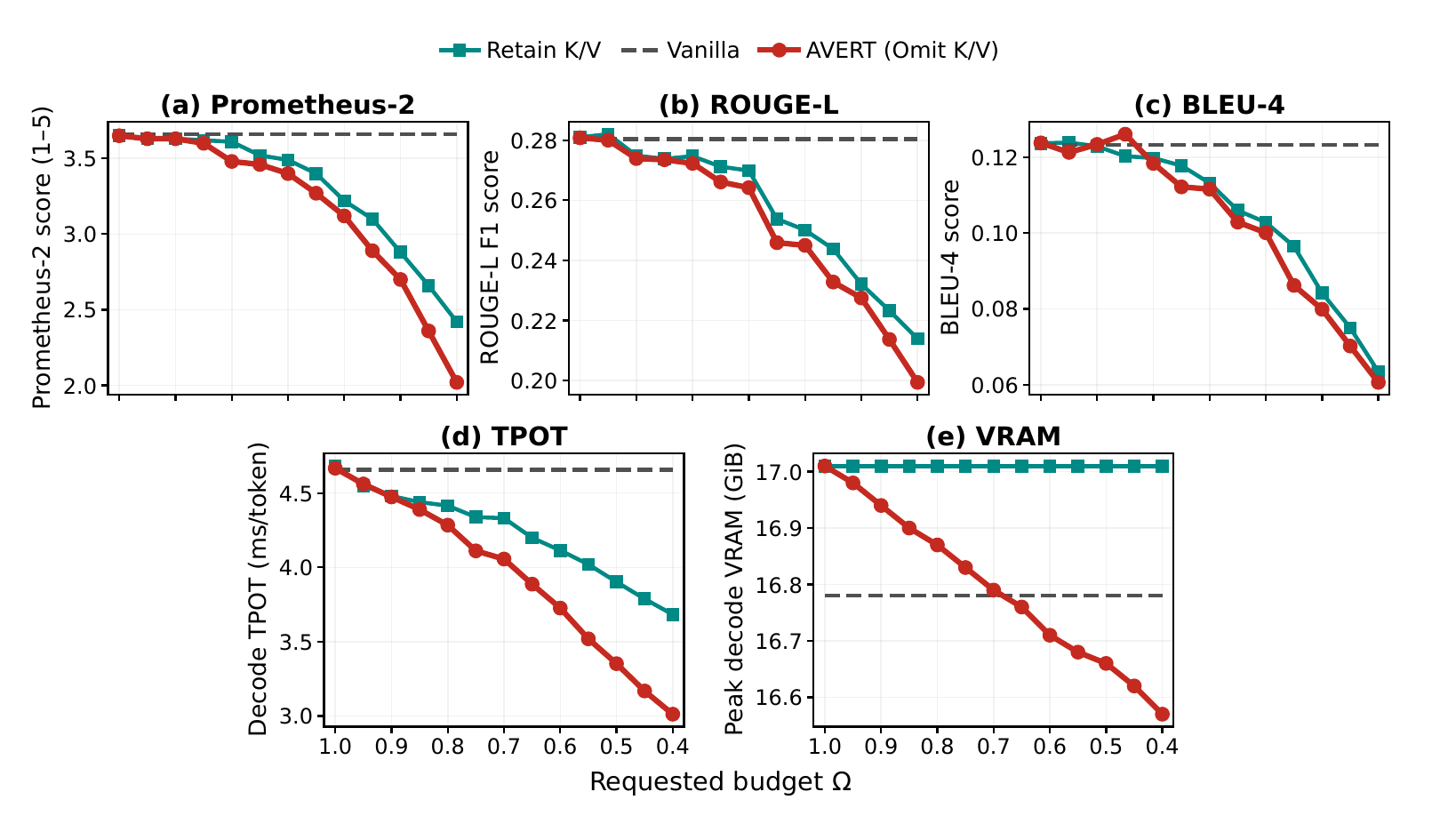}
\caption{\textbf{K/V omission versus retention on LLaMA-3.1-8B (GQA).} GQA already compresses K/V, so the memory gap is smaller than for MHA.}
\label{fig:app-s1s2-llama31}
\end{figure}

\begin{figure}[!htbp]
\centering
\includegraphics[width=0.94\linewidth]{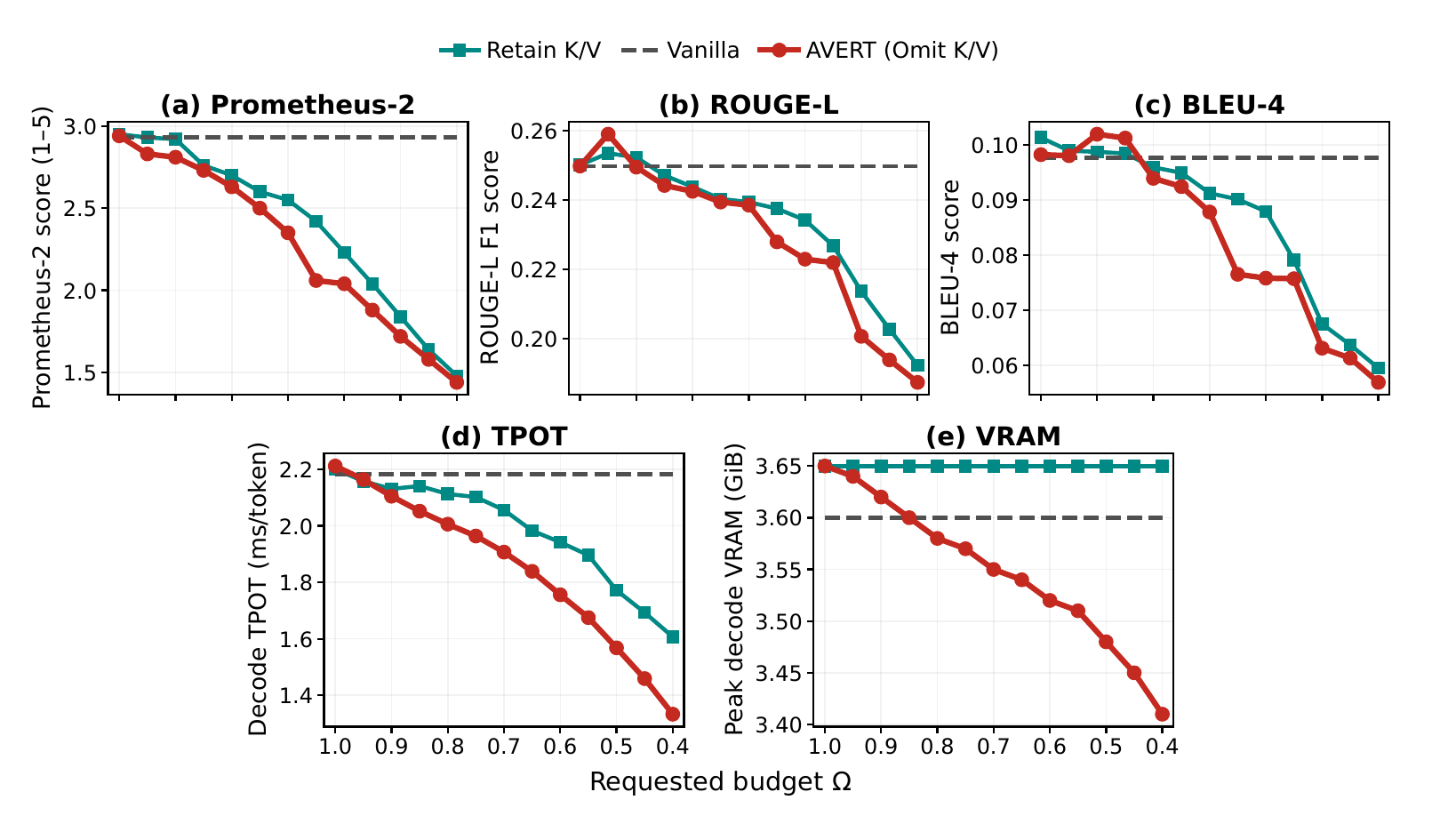}
\caption{\textbf{K/V omission versus retention on LLaMA-3.2-1B (GQA, 16L).}}
\label{fig:app-s1s2-llama32}
\end{figure}

Figures~\ref{fig:efficiency-grid}--\ref{fig:app-s1s2-llama32} compare Prometheus, ROUGE-L, BLEU-4, decode TPOT, and peak decode VRAM. In the retention ablation, a skipped token computes and stores its K/V projections and passes through SCA. It skips the attention output and FFN.

Retention keeps the full attention history and a nearly constant cache footprint. Omission lowers peak decode VRAM as the budget falls. The two paths have similar TPOT near $\omegaT=1.0$. As the budget falls, TPOT drops for both, with omission generally faster.

\FloatBarrier
\section{Batch-size sensitivity}
\label{app:bs-full}
We evaluate BS $\in\{1,2,4,8,16,32\}$ on the 1,000-example Alpaca set with LLaMA-3.1-8B. The three Alpaca quality metrics use the standard generation protocol. TPOT and peak decode VRAM are measured separately with a fixed 128-token prefill and 128 decode-produced tokens at every batch size. \method{} quality remains stable as batching changes, while TPOT falls as GPU utilization improves and peak decode VRAM grows with concurrent requests. Across the six tested batch sizes, \method{} varies by at most 0.15 on Prometheus-2, 0.0135 on ROUGE-L, and 0.0119 on BLEU-4. For GateSkip-FFT at requested $\omegaT=0.4$, Prometheus-2 varies by 1.99 points.

\begin{figure}[!htbp]
\centering
\includegraphics[width=0.96\linewidth]{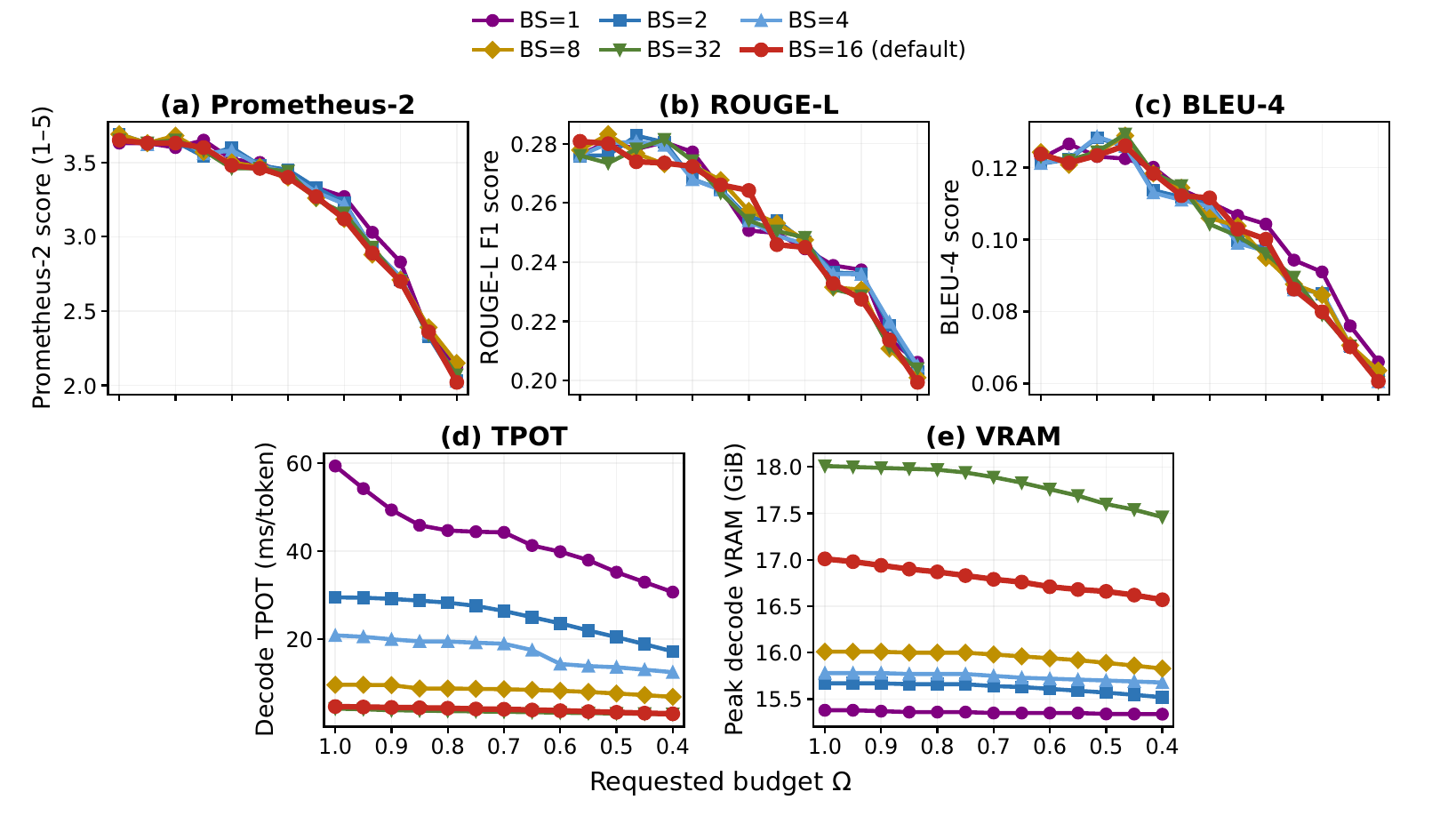}
\caption{\textbf{\method{} quality and decode efficiency by batch size on LLaMA-3.1-8B.}}
\label{fig:app-bs-all}
\end{figure}

\begin{figure}[!htbp]
\centering
\includegraphics[width=0.96\linewidth]{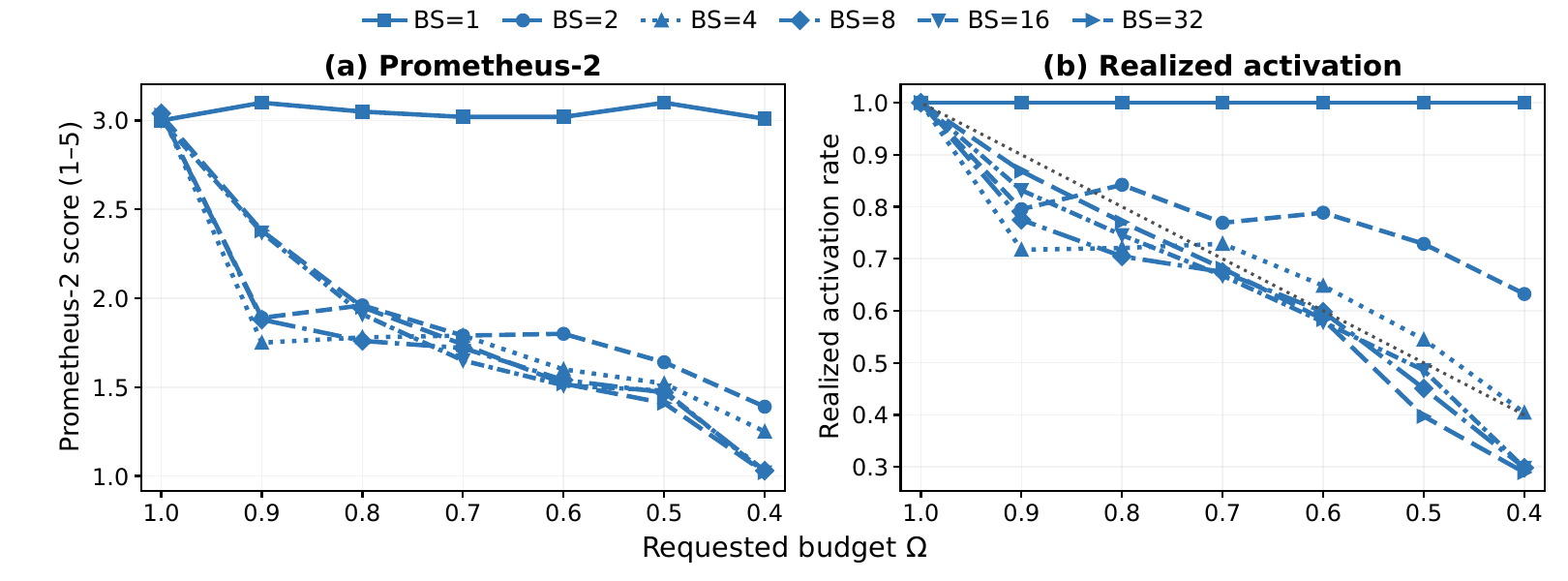}
\caption{\textbf{GateSkip-FFT batch-size sensitivity on LLaMA-3.1-8B.} Left: Prometheus versus requested $\omegaT$. Right: execution fraction versus requested $\omegaT$, with the identity line as the calibration target. Both quality and realized compute change sharply with BS because the selector uses a threshold derived from the current batch-wide live-token pool.}
\label{fig:app-gateskip-bs}
\end{figure}

\section{Per-request budget control and mixed batching}
\label{app:mixed-split}

\subsection{Token depth and budget error within a request}
\label{app:request-depth}
Using LLaMA-3.1-8B, we generate a response to each of 160 prompts at each of three budgets, giving 480 responses. The generation limit is 128 tokens. We record how many layers execute at each routed decode position, excluding the first two full-depth positions. For each response we average these counts, divide by 32, and take the absolute difference from its budget. Every response receives equal weight (Table~\ref{tab:rebuttal-request-error}).

\begin{table}[!htbp]
\centering
\caption{Token depth and budget error on LLaMA-3.1-8B. Errors are in percentage points (pp). The last column counts responses whose average execution fraction is within two points of their target.}
\label{tab:rebuttal-request-error}
\small
\begin{tabular}{@{}lrrrrr@{}}
\toprule
Budget & Responses & Routed positions & \makecell[r]{Mean absolute\\error (pp)} & \makecell[r]{95th percentile\\(pp)} & Within 2 pp \\
\midrule
0.40 & 160 & 17,547 & 1.150 & 2.000 & 153/160 \\
0.60 & 160 & 18,089 & 1.150 & 1.901 & 154/160 \\
0.80 & 160 & 17,984 & 1.150 & 1.785 & 153/160 \\
All  & 480 & 53,620 & 1.150 & 1.975 & 460/480 \\
\bottomrule
\end{tabular}
\end{table}

\subsection{Budget calibration in single- and mixed-budget batches}
We evaluate single-budget and mixed-budget batches at BS=32. For each request we take the absolute difference between its average execution fraction and its target. For each backbone, we average these errors separately at each budget and report the largest of these averages. For LLaMA-2-7B, LLaMA-3.1-8B, and LLaMA-3.2-1B, these maxima are 0.85, 0.97, and 0.94 percentage points for single-budget batches and 1.06, 1.15, and 0.89 points for mixed-budget batches. Each full mixed batch contains four requests at each of the eight targets 1.0, 0.9, 0.8, 0.7, 0.6, 0.5, 0.45, and 0.4.

\subsection{Mixed versus budget-separated execution}
Both schedules use the same checkpoint and the same 128 request--budget pairs. The four budgets are 1.0, 0.8, 0.6, and 0.4. In this order, the request counts are $(32,32,32,32)$ for the balanced workload, $(16,24,32,56)$ for moderate skew, $(8,16,24,80)$ for low-budget skew, and $(80,24,16,8)$ for high-budget skew. Each request has 128 prompt tokens and generates exactly 128 output tokens with EOS stopping disabled. Mixed execution fills four batches of 32. Grouping by budget gives 4, 5, 6, and 6 batches with mean fill rates of 100\%, 80\%, 66.7\%, and 66.7\%, respectively.

\begin{table}[!htbp]
\centering
\caption{\textbf{Mixed-budget versus homogeneous split execution.} Means over 10 runs using the same \method{} checkpoint and identical request--budget pairs. Batch counts shown as mixed/split; fill is the average split-batch fill ratio.}
\label{tab:app-mixed-split}
\scriptsize
\renewcommand{\arraystretch}{1.4}
\setlength{\tabcolsep}{3.0pt}
\resizebox{\textwidth}{!}{%
\begin{tabular}{ll cc cc cc}
\toprule
Workload & Model & \multicolumn{2}{c}{Total wall-clock (s)} & \multicolumn{2}{c}{Decode TPOT (ms/token)} & \multicolumn{2}{c}{Peak decode VRAM (GiB)} \\
\cmidrule(lr){3-4}\cmidrule(lr){5-6}\cmidrule(lr){7-8}
& & Mixed & Split & Mixed & Split & Mixed & Split \\
\midrule
\multirow{3}{*}{\makecell[l]{Balanced\\4/4 batches; 100\% fill}}
 & LLaMA-2-7B   & 54.53 & 54.47 & 3.263 & 3.258 & 17.13 & 19.01 \\
 & LLaMA-3.1-8B & 57.78 & 57.91 & 3.480 & 3.484 & 17.89 & 18.01 \\
 & LLaMA-3.2-1B & 29.61 & 29.65 & 1.780 & 1.781 & 3.47 & 4.03 \\
\midrule
\multirow{3}{*}{\makecell[l]{Moderate skew\\4/5 batches; 80\% fill}}
 & LLaMA-2-7B   & 50.28 & 53.72 & 3.003 & 3.214 & 17.04 & 18.13 \\
 & LLaMA-3.1-8B & 54.57 & 56.88 & 3.280 & 3.428 & 17.88 & 17.77 \\
 & LLaMA-3.2-1B & 27.08 & 28.34 & 1.625 & 1.702 & 3.47 & 3.65 \\
\midrule
\multirow{3}{*}{\makecell[l]{Strong low-budget skew\\4/6 batches; 66.7\% fill}}
 & LLaMA-2-7B   & 46.79 & 59.14 & 2.791 & 3.544 & 17.03 & 17.55 \\
 & LLaMA-3.1-8B & 52.15 & 60.60 & 3.133 & 3.643 & 17.89 & 17.60 \\
 & LLaMA-3.2-1B & 24.85 & 26.82 & 1.489 & 1.608 & 3.47 & 3.50 \\
\midrule
\multirow{3}{*}{\makecell[l]{Strong high-budget skew\\4/6 batches; 66.7\% fill}}
 & LLaMA-2-7B   & 61.72 & 77.15 & 3.702 & 4.643 & 19.00 & 19.01 \\
 & LLaMA-3.1-8B & 62.61 & 71.34 & 3.779 & 4.307 & 18.10 & 18.02 \\
 & LLaMA-3.2-1B & 32.59 & 34.86 & 1.961 & 2.100 & 4.04 & 4.04 \\
\bottomrule
\end{tabular}%
}
\end{table}

For each run we sum the batch times and take the largest per-batch peak decode VRAM. Table~\ref{tab:app-mixed-split} gives the means over ten runs. TPOT and total decode time fall by the same percentage because every workload produces 16,384 tokens. The split schedule includes homogeneous high-budget batches, so its peak memory can exceed that of a mixed batch. On LLaMA-3.1-8B, peak decode VRAM was 17.89 GiB for mixed batches and 18.01 GiB for split batches in the balanced workload. With mostly low-budget requests, the peaks were 17.89 and 17.60 GiB, respectively.

Each balanced mixed batch contains $(8,8,8,8)$ requests in the budget order above. With moderate skew, the four batches use $(8,8,8,8)$, $(8,8,8,8)$, $(0,8,12,12)$, and $(0,0,4,28)$ in sequence. With low-budget skew they use $(8,8,8,8)$, $(0,8,12,12)$, $(0,0,4,28)$, and $(0,0,0,32)$. With high-budget skew they use $(8,8,8,8)$, $(12,12,8,0)$, $(28,4,0,0)$, and $(32,0,0,0)$.

\section{Evaluation between the reported budget grid points}
\label{app:offgrid}
\begin{figure}[!htbp]
\centering
\includegraphics[width=0.96\linewidth]{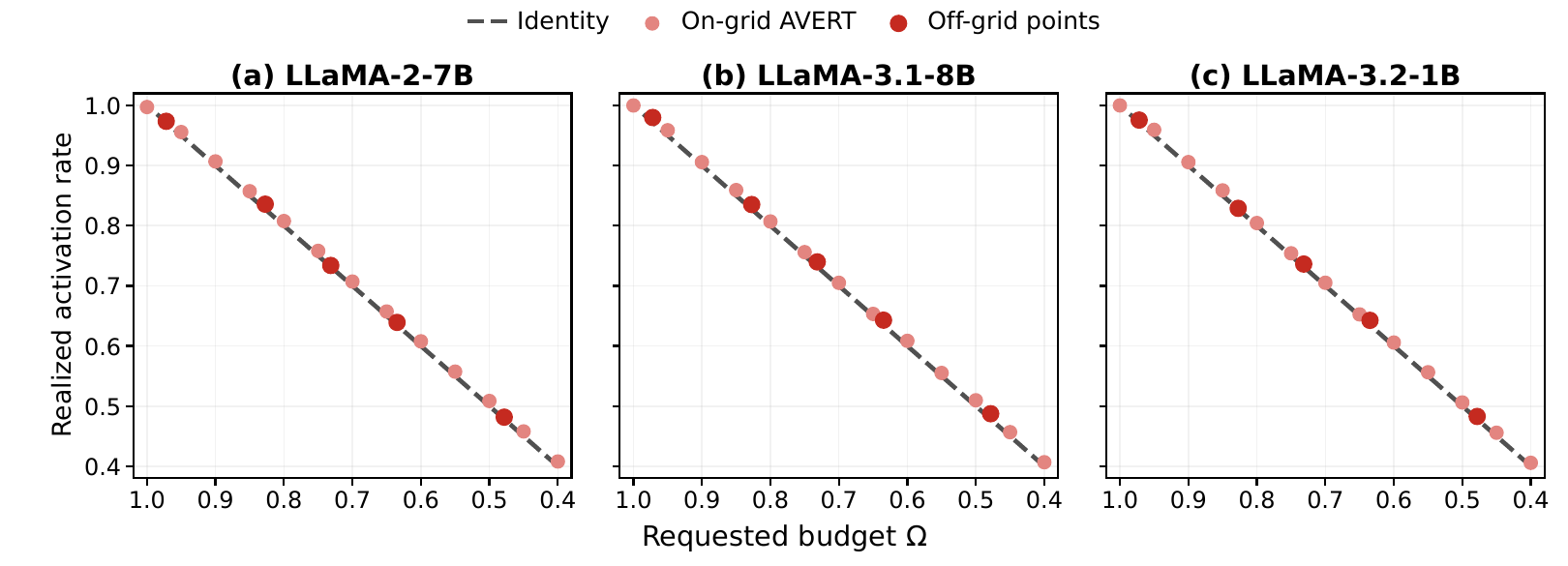}
\caption{\textbf{On-grid and off-grid budgets.} Requested budget versus average execution fraction. The diagonal marks equality of these quantities. The five off-grid targets are overlaid on the 13-point grid.}
\label{fig:offgrid-calibration}
\end{figure}

\begin{figure}[!htbp]
\centering
\includegraphics[width=0.96\linewidth]{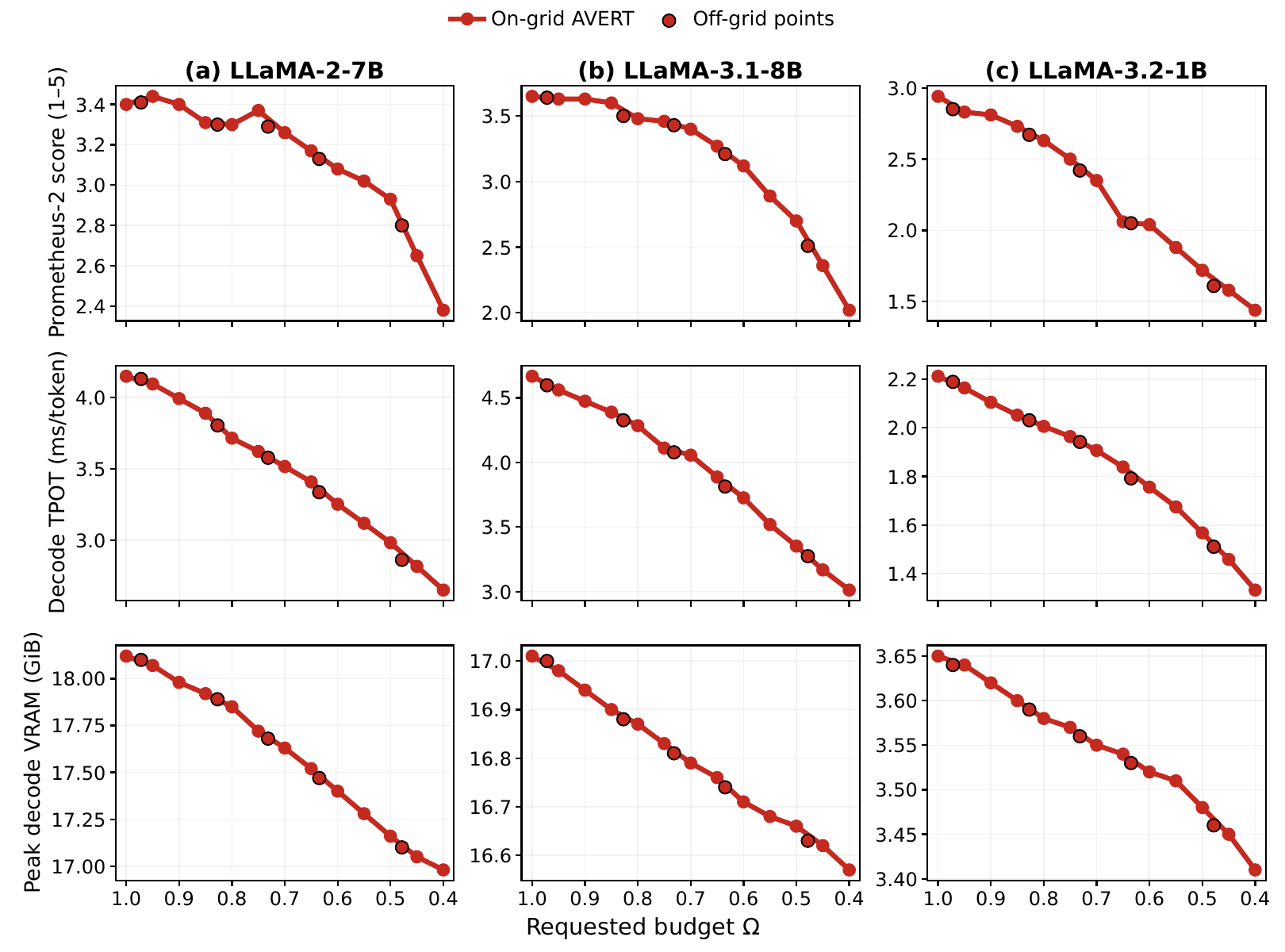}
\caption{\textbf{Off-grid operating points within the trained budget interval.} The 13 on-grid \method{} points define the requested-budget curves; the five off-grid evaluations are overlaid for Prometheus-2, decode TPOT, and peak decode VRAM.}
\label{fig:offgrid-continuity}
\end{figure}

We evaluate five targets that are not part of the reported 0.05-spaced grid, using the same values on all three backbones: $\omegaT\in\{0.4783,0.6348,0.7316,0.8273,0.9719\}$. For each backbone, we average the absolute budget errors of requests assigned the same target. Taking the largest of these averages over the five targets gives 0.85, 0.90, and 0.78 percentage points for LLaMA-2-7B, LLaMA-3.1-8B, and LLaMA-3.2-1B, respectively. Figure~\ref{fig:offgrid-calibration} compares these measurements with the 13 on-grid targets. Figure~\ref{fig:offgrid-continuity} overlays the same off-grid evaluations on the on-grid Prometheus-2, decode TPOT, and peak decode VRAM curves.

\FloatBarrier
\section{Matched-quality efficiency against AdaSkip}
\label{app:adaskip-efficiency}
We compare \method{} with AdaSkip at three pairs of configurations with similar Prometheus-2 scores for each backbone. The pairs were selected from the existing quality sweeps before TPOT and peak decode VRAM measurement; the absolute Prometheus difference is at most 0.06 in every pair. $\omegaT$ targets the average fraction of layers executed by \method{}, while AdaSkip's keep rate specifies the fraction of attention and FFN sublayers retained.

Both methods use the same timing protocol. We take the first 176 held-out Alpaca requests so every batch has BS=16. Batch 0 is a warm-up batch. Batches 1--10 contain the 160 timed requests; every request uses a fixed 128-token prefill and is forced to generate 128 output tokens, giving 20,480 timed output tokens in total. TPOT is the sum of decode time over those ten batches divided by $10\times16\times128$. Peak decode VRAM is measured in GiB with \texttt{torch.cuda.max\_memory\_allocated()/(1024**3)}. All measurements use the same NVIDIA A800 80GB GPU.

As for all compared methods, prefill executes every layer, and the selected attention/FFN skip set is applied only during decode.

At the AdaSkip keep rates used below, its fixed attention/FFN sublayer skip pattern gives lower TPOT and peak decode VRAM than \method{} at numerically equal keep rates. Under matched Prometheus-2 quality, AdaSkip must retain more sublayers. \method{} has lower TPOT in 7/9 pairs and lower peak decode VRAM in 3/9; in the remaining six memory comparisons, \method{} is higher by at most 5.7\%.

\begin{table}[!htbp]
\centering
\caption{\textbf{Prometheus-2-matched efficiency against AdaSkip.} Each row compares two operating points with similar Prometheus-2 quality. TPOT is in ms/token and peak decode VRAM in GiB.}
\label{tab:adaskip-matched}
\scriptsize
\setlength{\tabcolsep}{3.0pt}
\resizebox{\textwidth}{!}{%
\begin{tabular}{lcccc}
\toprule
Model & Keep rate (AVERT/AdaSkip) & Prometheus (AVERT/AdaSkip) & TPOT (AVERT/AdaSkip) & Peak decode VRAM (AVERT/AdaSkip) \\
\midrule
LLaMA-2-7B   & .80/.95 & 3.30/3.29 & 3.714/3.891 & 17.85/17.43 \\
LLaMA-2-7B   & .55/.90 & 3.02/3.02 & 3.119/3.678 & 17.28/16.88 \\
LLaMA-2-7B   & .45/.85 & 2.65/2.61 & 2.818/3.406 & 17.05/16.13 \\
\midrule
LLaMA-3.1-8B & .85/.95 & 3.60/3.56 & 4.389/4.419 & 16.90/16.51 \\
LLaMA-3.1-8B & .75/.90 & 3.46/3.44 & 4.111/4.217 & 16.83/16.23 \\
LLaMA-3.1-8B & .55/.85 & 2.89/2.85 & 3.519/3.942 & 16.68/15.86 \\
\midrule
LLaMA-3.2-1B & .70/.95 & 2.35/2.36 & 1.907/1.863 & 3.55/3.56 \\
LLaMA-3.2-1B & .55/.90 & 1.88/1.90 & 1.675/1.644 & 3.51/3.53 \\
LLaMA-3.2-1B & .40/.80 & 1.44/1.38 & 1.333/1.485 & 3.41/3.45 \\
\bottomrule
\end{tabular}%
}
\end{table}

\section{Full SCA ablation results}
\label{app:sca-full}
The no-SCA model is trained without SCA from the start and uses an identity skip path during both training and inference.

\begin{figure}[!htbp]
\centering
\includegraphics[width=0.96\linewidth]{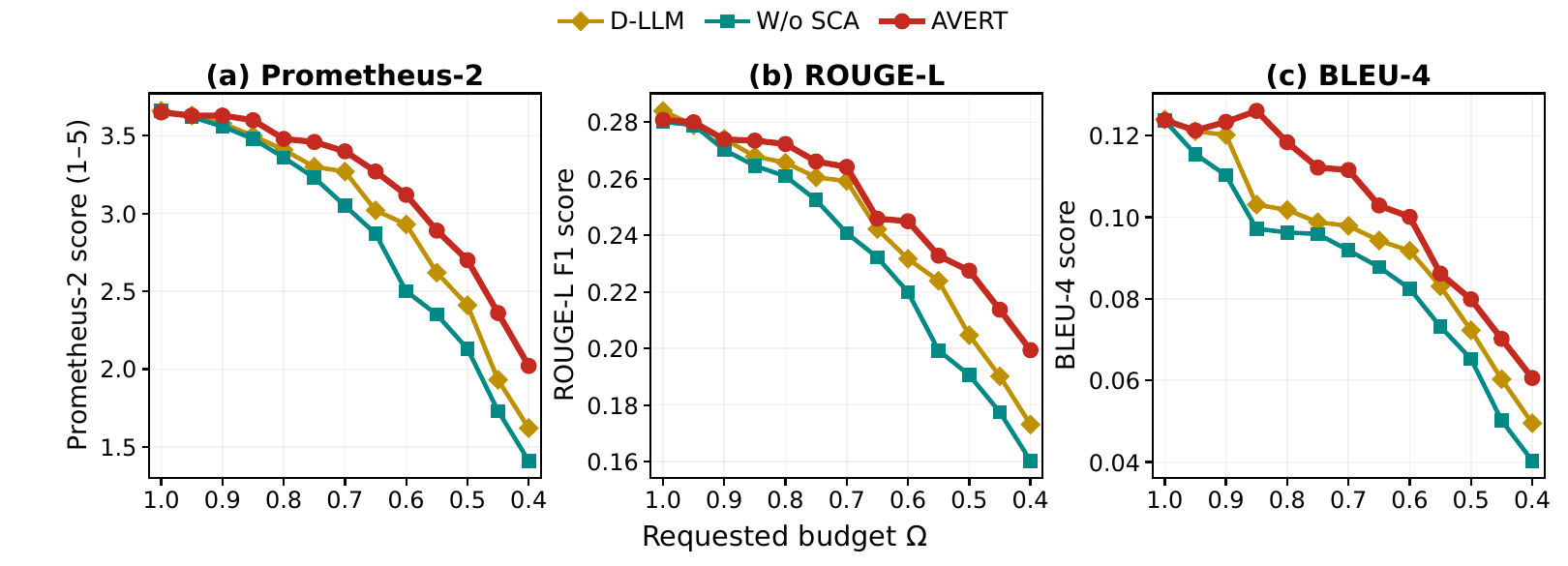}
\caption{\textbf{SCA ablation on LLaMA-3.1-8B.} The quality gap between full \method{} and the separately trained no-SCA model widens at lower requested budgets.}
\label{fig:app-sca}
\end{figure}

\section{Budget objectives and routing ablations}
\label{app:routing-objective-ablations}
All ablations in this appendix use LLaMA-3.1-8B.

\subsection{Request-average objective}
The per-token alternative uses the same positions eligible for routing, the same targets, and the same equal weighting across requests as Eq.~\ref{eq:active-loss}. It also counts all layers when calculating the execution fraction, but takes the absolute error before averaging over the tokens in each request. Deviations above and below $\omegaT_b$ therefore cannot offset one another, pushing each token toward the same depth. For fixed routes and the same request weights, the request-average loss is at most the per-token loss. For the two tokens executing 12 and 20 layers in Section~\ref{sec:depth-targets}, the losses are 0 and 0.125, respectively.

\subsection{Training losses}
\label{app:training-curves}
The two training runs use eight requests per micro-batch, three gradient accumulation steps, and the same learning-rate and budget-loss weight schedules. The request-average run optimizes $\mathcal L_{\mathrm{req}}$, logged as \texttt{a\_loss}; the per-token run optimizes $\mathcal L_{\mathrm{tok}}$, logged as \texttt{t\_loss}. Both also optimize answer-token cross-entropy (CE). Table~\ref{tab:rebuttal-training} reports the budget objective used for parameter updates in each run and its CE. 

\begin{table}[!htbp]
\centering
\caption{Training on LLaMA-3.1-8B. Entries are epoch means. The budget columns report the unweighted objectives used for parameter updates: request-average loss for \method{} and per-token loss for the alternative. CE is computed on supervised answer tokens.}
\label{tab:rebuttal-training}
\begin{tabular}{@{}r rr rr@{}}
\toprule
& \multicolumn{2}{c}{Request-average objective} & \multicolumn{2}{c}{Per-token objective} \\
\cmidrule(lr){2-3}\cmidrule(l){4-5}
Epoch & $\mathcal L_{\mathrm{req}}$ & Answer-token CE & $\mathcal L_{\mathrm{tok}}$ & Answer-token CE \\
\midrule
1 & 0.0708 & 1.2589 & 0.0880 & 1.3793 \\
2 & 0.0236 & 1.3702 & 0.0370 & 1.6000 \\
3 & 0.0176 & 1.2397 & 0.0290 & 1.4582 \\
4 & 0.0111 & 1.0707 & 0.0204 & 1.2401 \\
5 & 0.0070 & 0.9328 & 0.0164 & 1.0786 \\
\bottomrule
\end{tabular}
\end{table}

At epoch 5, $\lambda_{\mathrm{budget}}=10$, so the weighted budget terms are 0.070 and 0.164. They account for approximately 7.0\% and 13.2\% of the respective mean combined losses, calculated from the epoch-mean CE and weighted budget terms.

We also record request-average budget error for both runs as a common calibration metric. In the per-token run, this logged \texttt{a\_loss} is not used for parameter updates. Its epoch-wise means are 7.62, 2.60, 1.99, 1.19, and 0.77 percentage points. For the request-average run, the corresponding errors are 7.08, 2.36, 1.76, 1.11, and 0.70 points. Both runs first average below two points in epoch 3 and below one point in epoch 5.

\subsection{Progressive versus one-shot routing}
The one-shot ablation computes every dynamic-layer routing decision from the same hidden state at the start of the dynamic-layer stack. It retains the first two always-executed layers, the two full-execution decode positions, compact K/V omission, SCA, and the request-average objective. Progressive routing has higher ROUGE-L at six targets.

\begin{figure}[!htbp]
\centering
\includegraphics[width=0.96\linewidth]{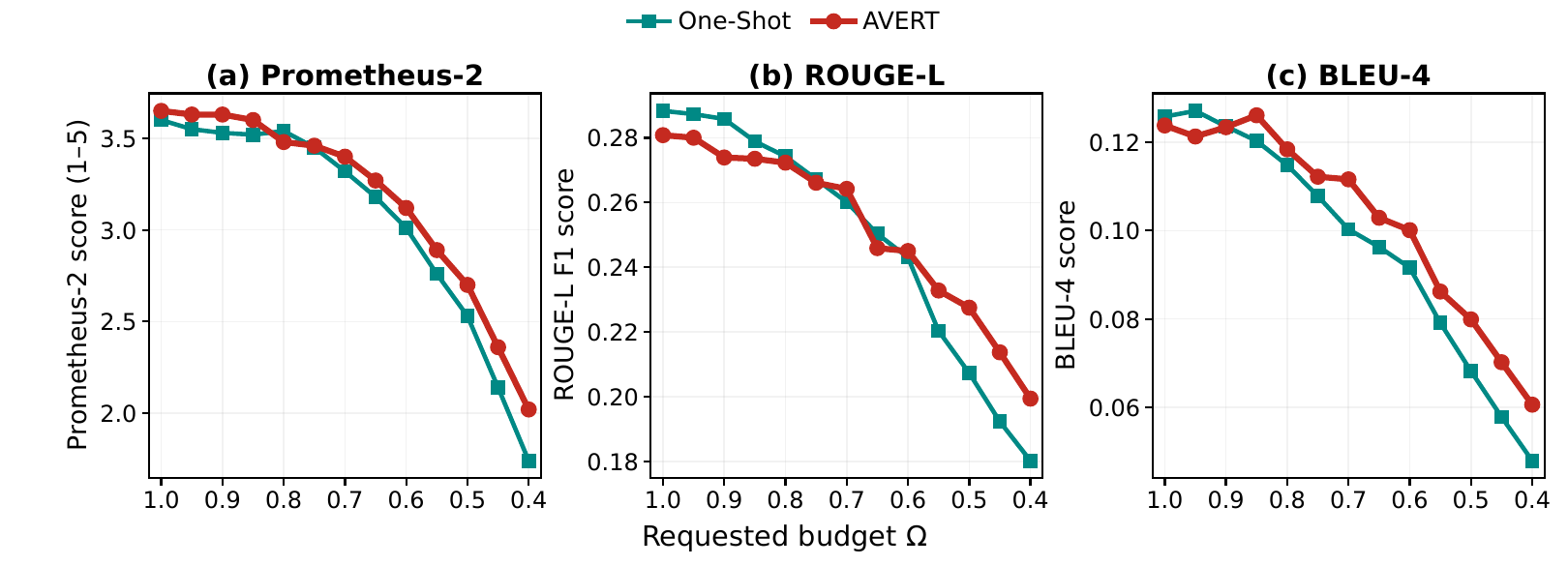}
\caption{\textbf{Progressive-routing ablation on LLaMA-3.1-8B.} One-shot routing decisions are all computed from the same hidden state before the dynamic-layer stack; \method{} makes each routing decision using the current hidden state at that layer.}
\label{fig:app-one-shot}
\end{figure}

\subsection{Sensitivity to the budget penalty}
\label{app:penalty}

\begin{figure}[!htbp]
\centering
\includegraphics[width=\textwidth]{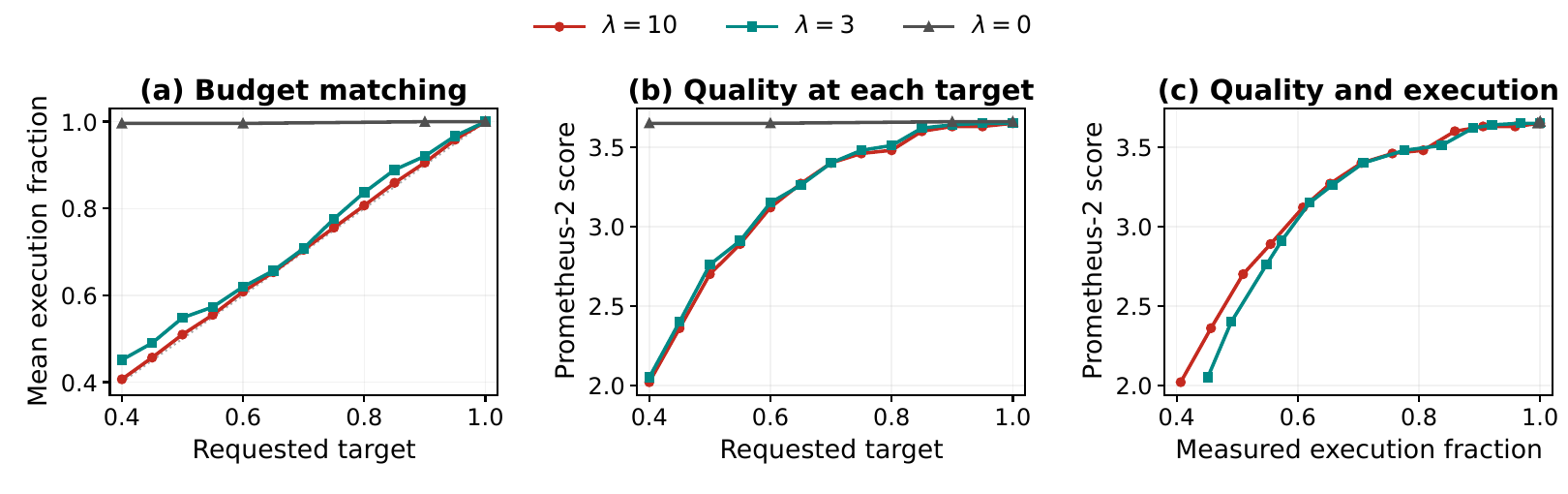}
\caption{\textbf{Penalty sensitivity on LLaMA-3.1-8B.} (a) Mean executed fraction versus requested target. (b) Prometheus-2 versus requested target. (c) Prometheus-2 versus the measured execution fraction. The $\lambda=0$ run includes target $1.0$ and three lower targets; its nearly coincident points reflect almost full execution.}
\label{fig:rebuttal-lambda}
\end{figure}

We compare the main \method{} checkpoint trained with $\lambda_{\mathrm{budget}}=10$ with runs trained using weights 3 and 0. Figure~\ref{fig:rebuttal-lambda} shows the requested budget, average execution fraction, and Prometheus-2. For each of the 13 targets, we compute the mean absolute request error (Section~\ref{sec:setup}). With $\lambda_{\mathrm{budget}}=10$, these 13 errors have a mean of 0.628 percentage points and a maximum of 0.970 points. With weight 3, the mean and maximum rise to 2.599 and 5.130 points.

At target 0.40, the run with weight 3 executes 45.13\% of layers and scores 2.05 on Prometheus-2. The standard model executes 40.67\% and scores 2.02. With weight 0, the model executes 99.60--100\% of layers at the four measured targets (Table~\ref{tab:rebuttal-lambda0}).

\begin{table}[!htbp]
\centering
\caption{\textbf{No-penalty control ($\lambda=0$).} Execution is the fraction of layers executed.}
\label{tab:rebuttal-lambda0}
\begin{tabular}{rrrrr}
\toprule
Target & Execution & Prometheus-2 & ROUGE-L & BLEU-4 \\
\midrule
1.00 & 0.999988 & 3.66 & 0.2813 & 0.1239 \\
0.90 & 0.999988 & 3.66 & 0.2813 & 0.1239 \\
0.60 & 0.996039 & 3.65 & 0.2812 & 0.1233 \\
0.40 & 0.996227 & 3.65 & 0.2809 & 0.1231 \\
\bottomrule
\end{tabular}
\end{table}

\FloatBarrier
\section{Parameter counts and execution overhead}
\label{app:overhead}
\subsection{Parameter counts}
Table~\ref{tab:param-overhead} gives parameter counts and storage at two bytes per parameter (fp16/bf16). Decode peaks also include the K/V cache and temporary allocations.

\begin{table}[!htbp]
\centering
\caption{\textbf{Static parameter overhead.} LoRA parameters are mergeable into the base model; router+SCA parameters remain at deployment.}
\label{tab:param-overhead}
\small
\begin{tabular}{lccc}
\toprule
Model & LoRA params (mergeable) & Router+SCA params & Non-mergeable overhead \\
\midrule
LLaMA-2-7B & 35.78M & 125.91M & $\approx$240 MiB \\
LLaMA-3.1-8B & 37.75M & 125.91M & $\approx$240 MiB \\
LLaMA-3.2-1B & 10.22M & 29.40M & $\approx$56 MiB \\
Qwen2.5-3B & 27.57M & 71.39M & $\approx$136 MiB \\
\bottomrule
\end{tabular}
\end{table}

\subsection{Component measurements}
\label{app:components}
\begin{figure}[!htbp]
\centering
\includegraphics[width=0.83\textwidth]{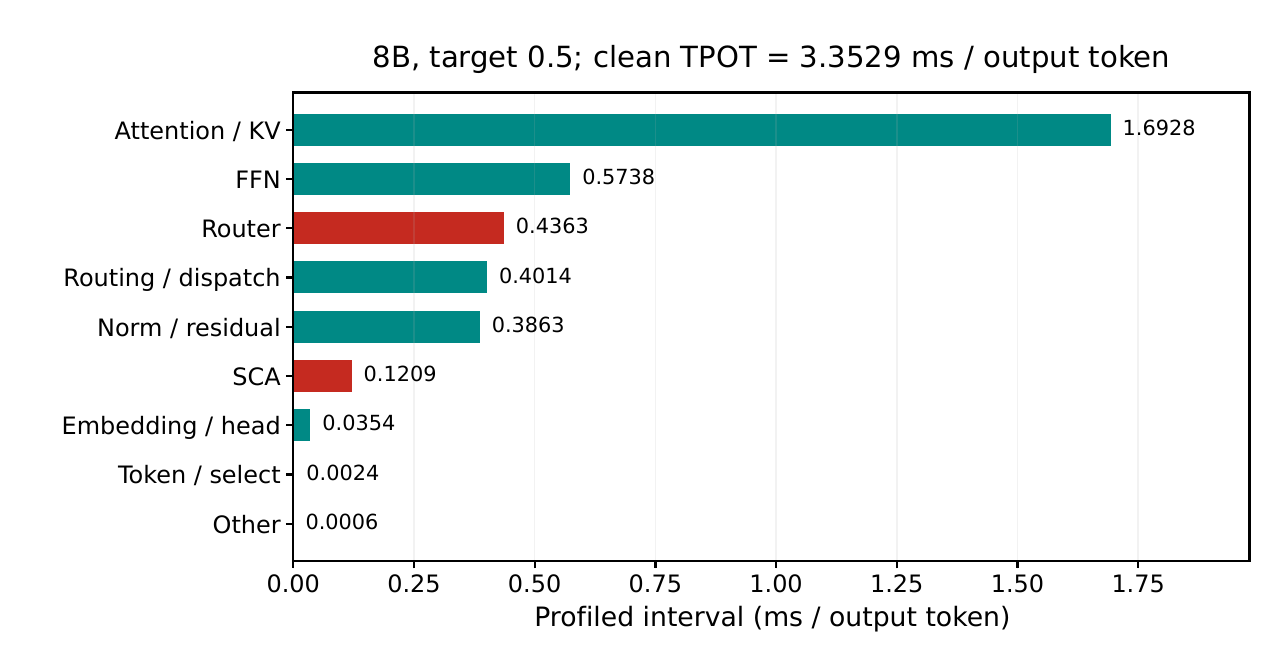}
\caption{\textbf{Decode component timings on LLaMA-3.1-8B at $\Omega=0.5$.} Bars show instrumented CUDA stream intervals per output token. They cannot be summed to recover the uninstrumented TPOT.}
\label{fig:rebuttal-components}
\end{figure}

We time LLaMA-3.1-8B at $\Omega=0.5$ on one A800, with 16 requests per batch, 128 prompt tokens and 128 output tokens per request. The uninstrumented decode takes 3.3529 ms per output token across the batch. Adding the profiler increases wall time. Figure~\ref{fig:rebuttal-components} gives the CUDA stream intervals from the instrumented run. Attention and K/V operations take 1.6928 ms per output token and the FFN takes 0.5738. The router takes 0.4363, routing dispatch (indexing and gather/scatter) 0.4014, and SCA 0.1209.

\section{Logit-Lens Probe of Local Routing Stability}
\label{app:logit-lens}
We use a logit-lens-style counterfactual probe to analyze whether \method{} skips locally stable layer--token pairs. The first two request-local decode positions advance the actual trajectory at full depth but are excluded from this analysis. At each subsequent dynamic-layer position of the generation using the model's own routing decisions, we branch the current hidden state into execute and skip+SCA outcomes. We project both through the final normalization and LM head, and measure $|\log(r_{\mathrm{skip}}+1)-\log(r_{\mathrm{execute}}+1)|$ for the token selected by the actual routed decode path, with ranks starting at one. Only the router-selected branch advances and writes persistent K/V when it executes. This local current-token probe does not measure how omitted K/V affects future tokens.

\begin{figure}[!htbp]
\centering
\includegraphics[width=0.91\textwidth]{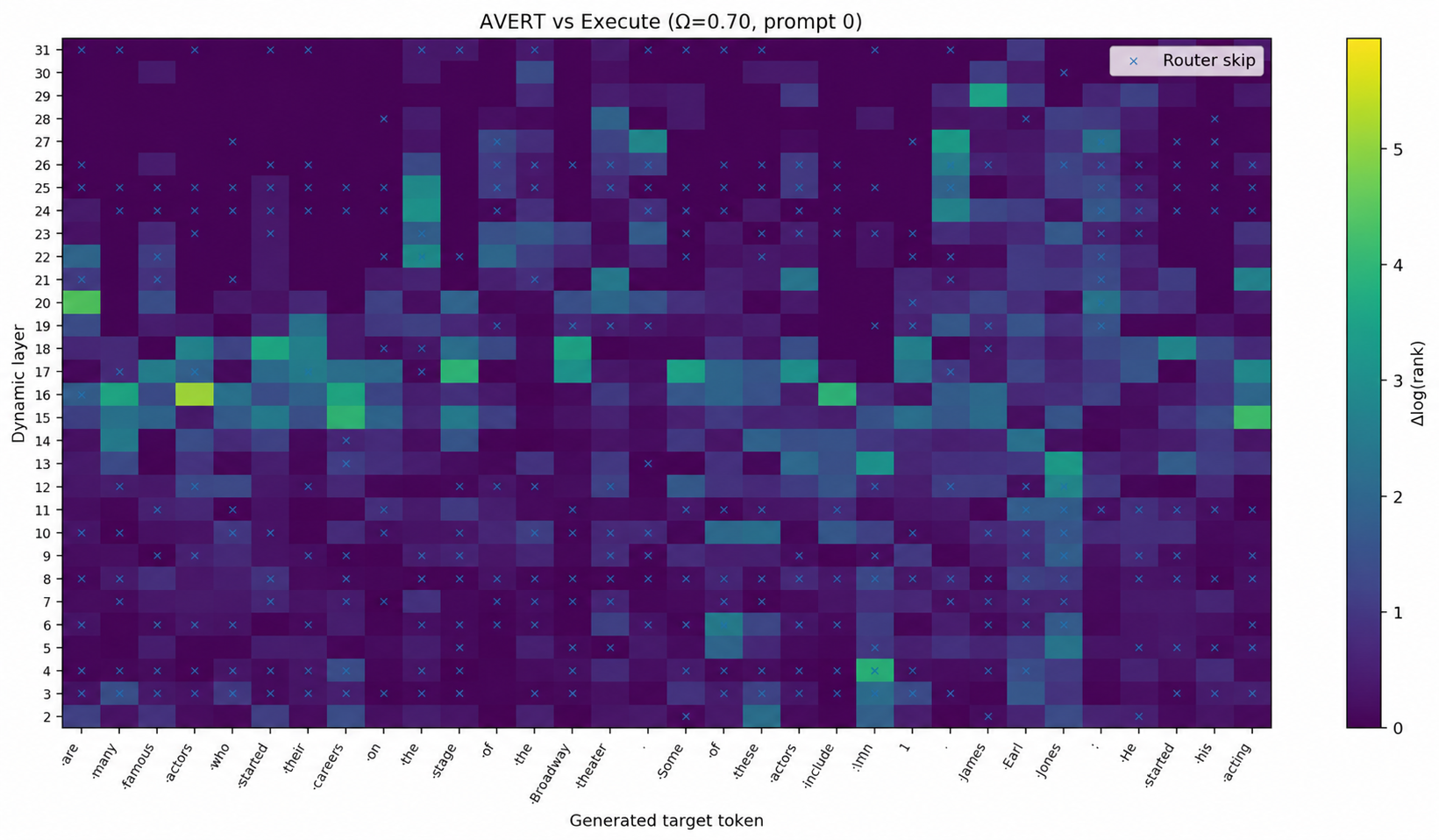}
\caption{\textbf{Qualitative logit-lens probe on LLaMA-3.1-8B at $\omegaT=0.70$.} Color indicates target-token $\Delta\log(\mathrm{rank})$ between skip and execute paths; crosses mark router-skipped pairs. Rows begin at zero-indexed layer 2 because the first two Transformer layers always execute and are not plotted. The first two full-execution decode inputs are also excluded.}
\label{fig:logit-lens-heatmap}
\end{figure}

\begin{figure}[!htbp]
\centering
\includegraphics[width=0.94\textwidth]{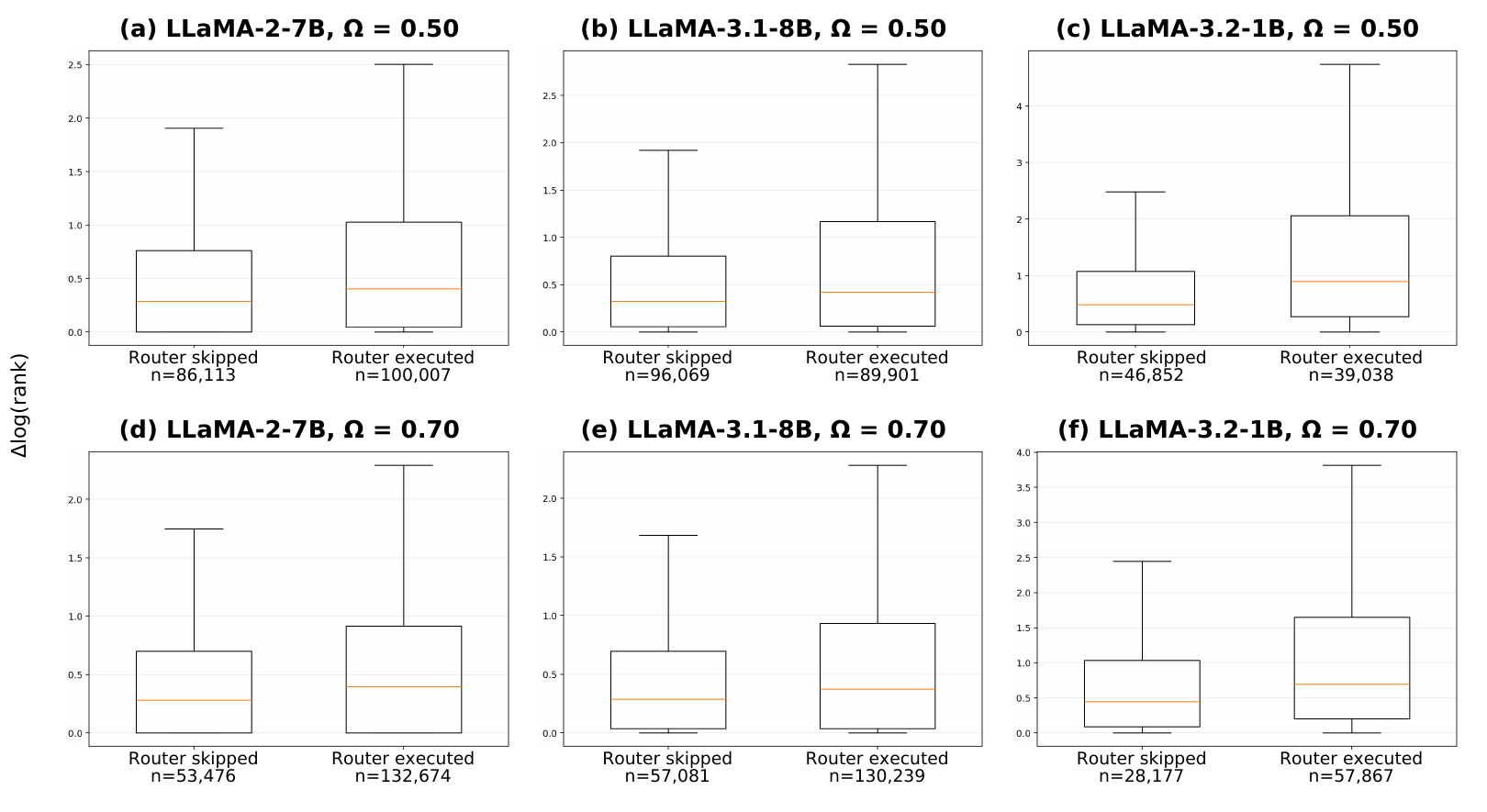}
\caption{\textbf{Aggregate local rank drift by router decision.} Across all three architectures and two targets, router-skipped layer--token pairs show distributions of local skip-versus-execute target-rank drift concentrated toward lower values than executed pairs.}
\label{fig:logit-lens-boxplots}
\end{figure}

\FloatBarrier
\section{Reproducibility Details}
\label{app:reproducibility}

\paragraph{Training data and optimization.}
All three main LLaMA \method{} models use the 51,564-example Alpaca training split included in the supplementary archive. The disjoint 1,000-example evaluation set contains the 805-example D-LLM test set plus 195 additional held-out Alpaca-format examples. We trained for five epochs with eight requests per micro-batch, three gradient accumulation steps, a maximum sequence length of 512, and seed 0. The effective batch size is \texttt{batch\_size}$\times$\texttt{accum\_iter}$\times$world size. Optimization uses AdamW with $\beta_1=0.9$, $\beta_2=0.95$, weight decay 0.05, and base learning-rate parameter $5\times10^{-3}$; the actual learning rate is scaled by effective batch size divided by 256. The schedule uses one epoch of linear warm-up followed by half-cycle cosine decay to zero. The budget-loss weight increases linearly from 0 to 10 over the first epoch and stays at 10 thereafter. The base model is frozen; LoRA, router, and SCA parameters are trained, with LoRA rank 16, router hidden width 512, SCA bottleneck 256, and dynamic routing beginning after the first two Transformer layers. LoRA is applied to the query, key, and value projections $W_q,W_k,W_v$ and the FFN up, down, and gate projections $W_{\mathrm{up}},W_{\mathrm{down}},W_{\mathrm{gate}}$; it is not applied to the attention output projection $W_o$. It uses dropout 0.05 and no additional $\alpha/r$ multiplier. The Vanilla references use rank-16 LoRA with the same dropout and Q/K/V and FFN projection scope, without routers or SCA. For \method{}, the frozen base weights are loaded in bfloat16, while the trainable LoRA, router, and SCA parameters are maintained in float32 during training.

\paragraph{Budget sampling.}
Each request is assigned a training target $\omegaT_b\in[0.4,1.0]$. For 90\% of training batches, request targets are sampled continuously and independently from the uniform distribution on this interval. For the remaining 10\%, one endpoint, $0.4$ or $1.0$, is selected at random and assigned to the current distributed microbatch. In distributed training, the sampled global target vector is broadcast from rank 0 before each worker selects its local slice. The default training path keeps prompt/prefill positions at full compute and applies the budget loss only where decode routing is active; the auxiliary PIQA variant enables routing on prefill positions as described in Section~\ref{sec:setup}. In the default decode-routing path, the first two decode positions after prefill are reserved for full execution and excluded from the budget loss. In the auxiliary PIQA path, the first two valid prefill positions are likewise forced to execute and excluded from its objective.

\paragraph{Hardware.}
The main LLaMA training, inference, quality evaluation, TPOT, and peak decode VRAM experiments use a single NVIDIA A800 80GB GPU. Eight requests per micro-batch and three accumulation steps give an effective batch size of 24 and a peak learning rate of $4.6875\times10^{-4}$. The released scripts use single-GPU execution with \texttt{NPROC\_PER\_NODE=1}. Core software dependencies are listed in the supplementary \texttt{requirements.txt}; the README records the optional vLLM requirement for Prometheus-2 scoring.

\paragraph{Baseline execution paths.}
D-LLM trains one checkpoint per target, requiring 13 training runs for our sweep. In its released implementation, later-token attention is computed over the full materialized history before the eviction mask is applied, so masked positions neither shorten attention nor reduce decode-cache storage. Our system measurements preserve its checkpoint and layer decisions but physically omit masked K/V entries. GateSkip-FFT uses the original full-parameter adaptation, while GateSkip-LoRA matches our LoRA setting. The released Hugging Face gate runs after Transformer computation; the authors' physical-skipping patch does not support cached autoregressive attention. GateSkip-FFT TPOT therefore measures its released cached-autoregressive path.

\paragraph{Evaluation configuration.}
The reported quality sweep uses 13 requested targets from 0.40 to 1.00 in increments of 0.05. Alpaca generation uses the full 1,000-example held-out split, batch size 16, maximum generation length 256, temperature 0, and top-$p=1$. GSM8K uses all 1,319 test questions with 8-shot prompting, batch size 16, maximum generation length 256, greedy decoding, and flexible answer extraction. IFEval uses all 541 official prompts in the zero-shot setting, batch size 16, maximum generation length 512, temperature 0, and top-$p=1$. The official prompt is wrapped in the Alpaca instruction template for generation; the bundled rule-based evaluator still receives the original official prompt and instruction keys. PIQA uses all 1,838 examples as an auxiliary prefill-skipping probe. The probe computes dense projections before applying its skip mask. Each example is formatted as \texttt{Question: <goal>} followed by \texttt{Answer:}; both candidate continuations are scored by log likelihood, and the reported accuracy selects the larger log likelihood after normalization by the candidate's UTF-8 byte length. The auxiliary checkpoint is trained only on the same 51,564-example Alpaca split; PIQA is used only for evaluation. Prometheus scoring uses \texttt{prometheus-eval/prometheus-7b-v2.0} in direct-assessment mode with bfloat16 inference, temperature 0, and a 512-token judge-output limit. The judge receives the instruction, generated response, reference answer, and a fixed rubric: scores 1--2 denote unusable or substantially incorrect responses, 3 denotes an acceptable response with errors or omissions, 4 denotes a mostly correct and complete response with minor issues, and 5 denotes an accurate, complete, fluent, and well-organized response. The judge returns feedback and an integer score. Malformed or unparseable judge outputs receive score 1; the released script records and reports their count. All evaluated models use bfloat16 inference for the reported generation, TPOT, and peak decode VRAM measurements. Unless otherwise stated in the paper, generation and K/V measurements use batch size 16; mixed-budget and batch-size experiments use the configurations specified in Sections~\ref{sec:exp-batching}--\ref{sec:exp-ablation} and Appendices~\ref{app:bs-full}--\ref{app:mixed-split}.

\paragraph{BUDDY reimplementation.}
No official BUDDY source code was publicly available when the comparison was prepared. Our explicit-budget reimplementation follows the paper-specified first and last always-executed Transformer blocks, Linear--SiLU--Linear Decision Module, budget-to-$k=\mathrm{round}(bL)-2$ mapping over the middle blocks, hard Top-$k$ selection, rerouting at every autoregressive decode step, first-layer Value states as the planner feature, and aligned zero-filled K/V positions for skipped states. The paper names a Token-Pooling operator and a Taylor prior without releasing their exact implementations.

We reconstruct Token-Pooling as a causal mean over first-layer Value states and use a first-order Taylor-based residual-removal proxy for the prior. The prior is added to the routing scores with weight 0.1 before Top-$k$ selection. The LoRA rank follows the setting of the original BUDDY paper.

For this comparison, BUDDY is trained for five epochs rather than the paper's two. Each request draws its training budget independently and uniformly from $[0.4,1.0]$. Training uses AdamW with learning rate $10^{-4}$, 100 warm-up updates, and effective batch size 24. Prompt positions are excluded from the language-model loss. The reported evaluations use full-compute prefill and apply layer skipping during decode. The optional Group Relative Policy Optimization (GRPO) budget predictor is not used because every BUDDY evaluation supplies an explicit budget.

At $\omegaT=1$, BUDDY executes all layers with LoRA weights trained across depths. Vanilla uses LoRA weights trained at full depth. On LLaMA-3.1-8B, their respective Prometheus-2 scores are 3.29 and 3.66, ROUGE-L scores are 0.298 and 0.280, and BLEU-4 scores are 0.126 and 0.123.

\paragraph{Synchronized TPOT measurement.}
The synchronized PyTorch workloads use fixed requests and exclude stochastic arrivals, queueing, and online continuous batching.
For all synchronized system benchmarks, including Qwen, prompts are left-truncated or attended left-padded to 128 positions unless the experiment specifies another physical prompt length. The full batch completes full-compute prefill before 128 single-token decode steps. The token produced by prefill bootstraps the first decode step and is excluded from both decode timing and the output count; EOS stopping is disabled. The prompt-length study instead uses physical prefill lengths from 128 to 2048 while keeping 128 decode-produced tokens. The LLaMA-3.1-8B batch-size sweep uses the same fixed-shape system workload, while its Alpaca quality metrics use the standard generation protocol.

For the standard BS=16 operating-point benchmark, one warm-up batch is followed by ten timed batches. TPOT is the synchronized decode wall-clock time over the ten timed batches divided by $10\times16\times128$. Timing includes model forward passes, routing, gather/scatter, SCA, compact-cache handling, greedy token selection, and the Python decode loop.

For \method{} and D-LLM, all layer weights remain resident during decoding. For AdaSkip, we determine the skipped attention/FFN sublayers before constructing the decode model. After full-compute prefill, we reload the model for decoding without instantiating those sublayers, reducing the weights held in GPU memory. After each batch's full prefill and bootstrap, prefill logits are released and CUDA peak statistics are reset. Peak decode VRAM is then measured over decode, and the operating-point value is the maximum per-batch decode peak across the warm-up and timed batches. \method{}'s per-request, per-layer K/V buffers round the initial required length to a 64-token block and leave one additional 64-token block of headroom; when a buffer fills, capacity grows to the rounded maximum of twice the current capacity and the required length. This budget-independent policy makes physical allocation stepwise rather than equal to the number of valid cached positions.

\section{Efficiency versus execution fraction}
\label{app:realized-activation-efficiency}
For \method{}, D-LLM, GateSkip-FFT, and GateSkip-LoRA, we record execution fraction at each method's routing granularity together with decode TPOT and peak decode VRAM under the fixed 128-token prefill, 128-token decode, BS=16 system protocol. For \method{}, the measured execution fraction excludes its two reserved full-execution decode positions. BUDDY is omitted from this diagnostic because its explicit-budget route executes an exact integer Top-$k$ count at every decode step, up to budget discretization.

\begin{figure}[!htbp]
\centering
\includegraphics[width=0.96\linewidth]{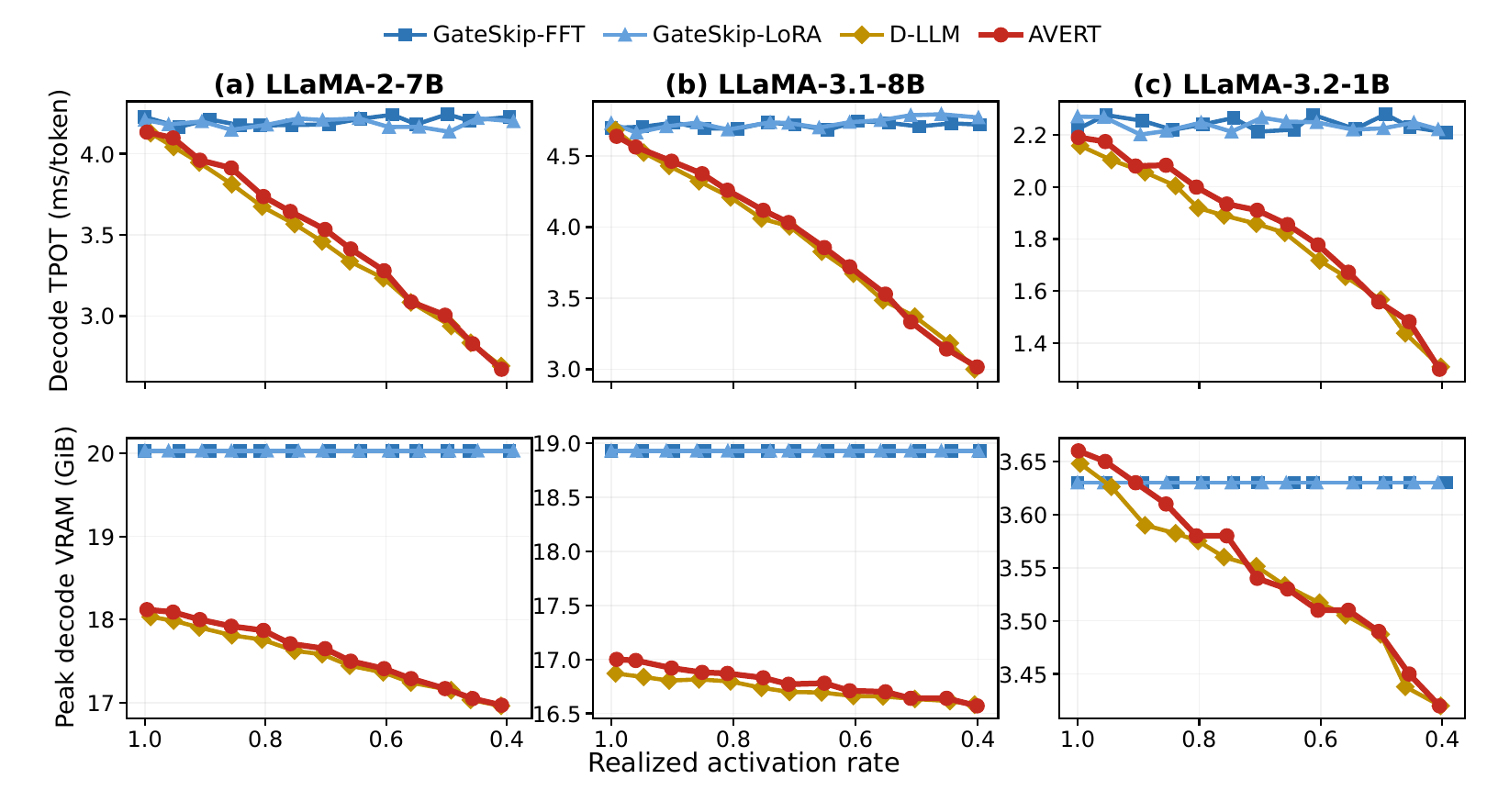}
\caption{\textbf{Decode efficiency versus measured execution fraction.} The top row shows decode TPOT and the bottom row shows peak decode VRAM. Each curve uses the execution fractions measured for that method at the 13 requested budgets.}
\label{fig:realized-activation-efficiency}
\end{figure}

\section{Existing quality--efficiency operating points}
\label{app:existing-quality-cost}
We join the measured Alpaca Prometheus scores with TPOT and peak decode VRAM measurements by backbone, method, and requested target. These are configuration-level comparisons: quality uses the generation evaluation protocol, while system measurements use the fixed 128-token prefill and 128-token output protocol. For each of the 195 baseline points (five baselines, three backbones, 13 targets), we select the evaluated \method{} point on the same backbone with the closest Prometheus-2 score. If two scores are equally close, we choose the higher \method{} budget. We keep pairs whose scores differ by at most 0.05. The supplementary CSV records all matched and unmatched anchors.

\begin{figure}[!htbp]
\centering
\includegraphics[width=0.96\linewidth]{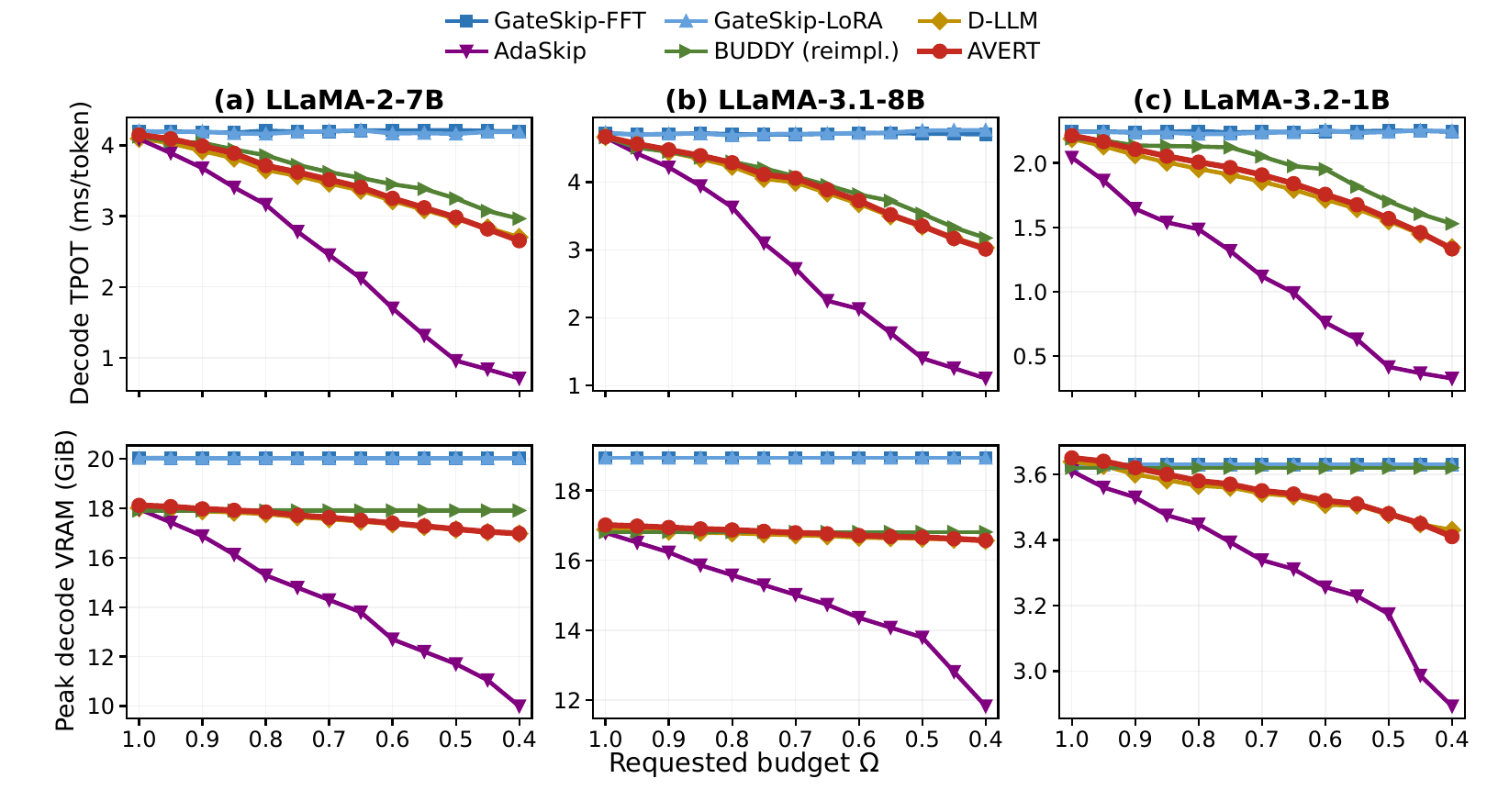}
\caption{\textbf{Measured TPOT and peak decode VRAM across requested targets.} GateSkip-FFT and merged GateSkip-LoRA are measured separately. }
\label{fig:app-tpot-vram}
\end{figure}

\begin{table}[!htbp]
\centering
\caption{Counts of matched pairs whose Prometheus-2 scores differ by at most 0.05. Several baseline points may match the same \method{} point.}
\label{tab:existing-quality-cost}
\small
\begin{tabular}{lccc}
\toprule
Baseline & Matched / 39 & Lower TPOT & Lower peak decode VRAM \\
\midrule
GateSkip-FFT & 10 & 10 & 10 \\
GateSkip-LoRA & 15 & 15 & 15 \\
D-LLM & 25 & 20 & 13 \\
AdaSkip & 11 & 7 & 3 \\
BUDDY (reimpl.) & 15 & 15 & 15 \\
\bottomrule
\end{tabular}
\end{table}
\method{} has lower peak decode VRAM in 13/25 matched D-LLM pairs and 3/11 matched AdaSkip pairs.

\FloatBarrier
\section{Token-depth allocation and layer sets across budgets}
\label{app:routing-behavior}
\begin{figure}[!htbp]
\centering
\includegraphics[width=0.96\linewidth]{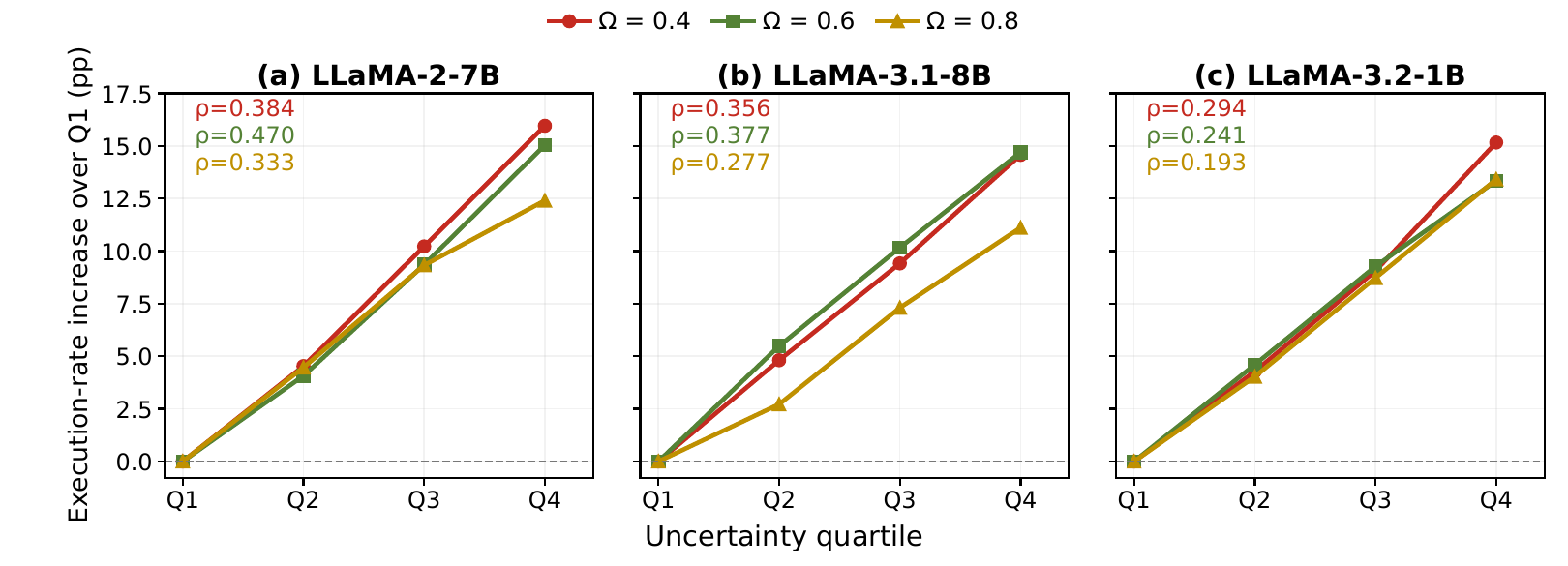}
\caption{\textbf{Token uncertainty and execution depth.} For each backbone and budget, the plot shows the mean fraction of layers executed in each entropy quartile relative to the lowest-entropy quartile (Q1), in percentage points. We compute Spearman's $\rho$ across individual tokens using dense-reference next-token entropy and executed depth.}
\label{fig:uncertainty-depth}
\end{figure}

\paragraph{Uncertainty and executed depth.}
For each requested target, we generate an \method{} continuation and replay its prefixes with the same checkpoint forced to execute all layers to obtain the entropy of the next-token prediction. The depth used to process a generated token is paired with the entropy of its next-token prediction; the first two full-execution decode positions and a final position without a successor target are excluded. Entropy is an offline difficulty proxy, not a router input. Figure~\ref{fig:uncertainty-depth} groups positions by entropy quartile and reports differences in the fraction of layers executed relative to Q1 in percentage points. Higher-uncertainty groups use greater mean depth in the displayed settings. Tokens are grouped within each target because the generated continuations can change with the budget. Under BUDDY's fixed explicit target, the integer Top-$k$ count is constant across tokens even when the selected layer identities change.

\begin{figure}[!htbp]
\centering
\includegraphics[width=0.96\linewidth]{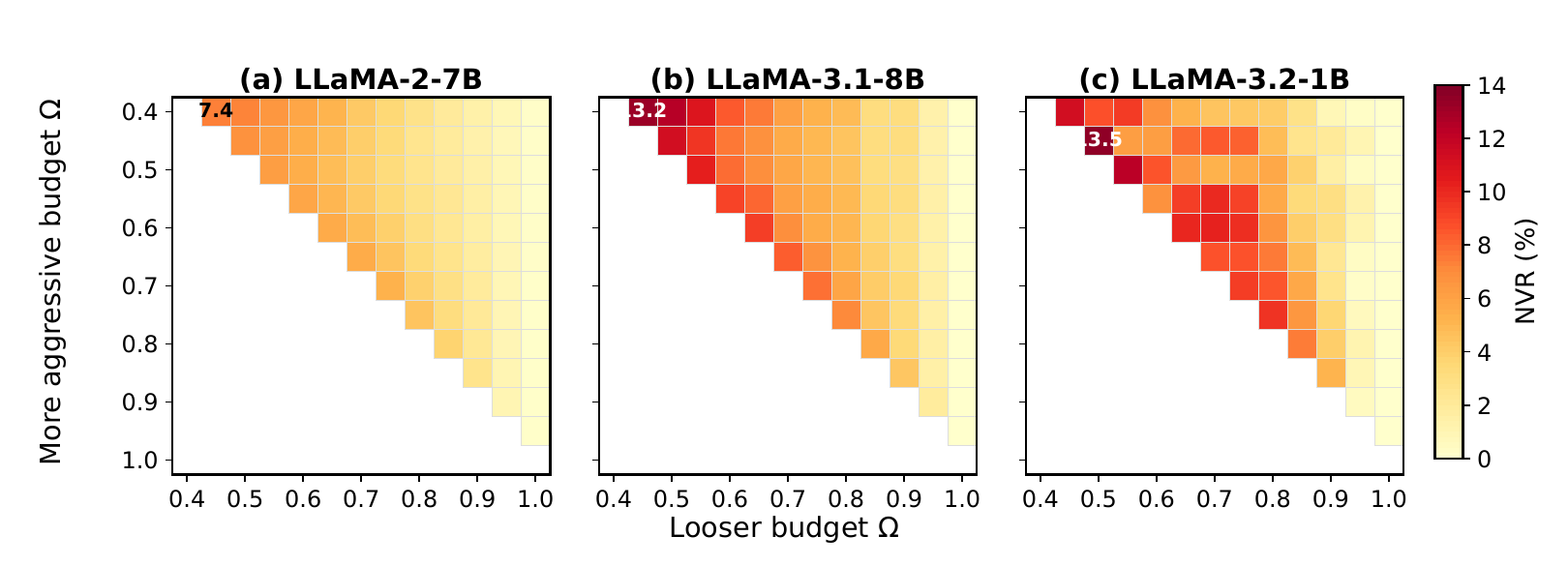}
\caption{\textbf{Non-nested selections across requested targets.} Each cell compares the same reference text under two budgets. Nonzero NVR means that some dynamically executed low-budget layer--token pairs are absent at the higher budget.}
\label{fig:budget-nesting}
\end{figure}

\paragraph{Non-nested layer selection.}
We teacher-force the same Alpaca reference continuations across all requested targets and record the executed sets $S_{\Omega,i}$ at each routed position $i$, excluding both the first two full-execution decode positions and the first two always-executed layers. For $\Omega_{\mathrm{low}}<\Omega_{\mathrm{high}}$, the nesting-violation rate is
\begin{equation}
 \mathrm{NVR}=\frac{\sum_i |S_{\Omega_{\mathrm{low}},i}\setminus S_{\Omega_{\mathrm{high}},i}|}{\sum_i |S_{\Omega_{\mathrm{low}},i}|}.
\end{equation}
We pool counts across positions before division. Figure~\ref{fig:budget-nesting} displays the 78 budget pairs from the 13-point grid. A positive NVR means that a layer executed at the lower budget is skipped at the higher budget for the same text position. The text is fixed, but hidden states and layer-local cache histories can change with routing.

\FloatBarrier
\section{Reduced Qwen2.5-3B evaluation}
\label{app:qwen25}
We evaluate Qwen2.5-3B \emph{base} (36 Transformer layers) at requested targets $\omegaT\in\{1.0,0.7,0.4\}$, comparing \method{}, our BUDDY reimplementation, and a separately finetuned Vanilla reference. This reduced sweep supplements the main LLaMA experiments and is excluded from their aggregate win counts. Generation-quality and system measurements use full-compute prefill. PIQA remains an auxiliary prefill-skipping probe; its \method{} results use the separately trained prefill-skipping checkpoint. Qwen2.5-3B is converted to the shared execution backend while retaining its native tokenizer; the supplementary code includes the converter and a parity check.

\begin{table}[!htbp]
\centering
\caption{\textbf{Qwen2.5-3B quality.} Prometheus is on the 1--5 scale; other metrics are fractions. SP/SI denote strict prompt/instruction accuracy. PIQA is the auxiliary prefill probe.}
\label{tab:qwen25-quality}
\scriptsize
\setlength{\tabcolsep}{3pt}
\resizebox{\textwidth}{!}{%
\begin{tabular}{lcccccccc}
\toprule
Method & $\omegaT$ & Prometheus & ROUGE-L & BLEU-4 & GSM8K-Flex & IFEval-SP & IFEval-SI & PIQA \\
\midrule
Vanilla & 1.00 & 3.460 & 0.2943 & 0.1263 & 0.5747 & 0.2680 & 0.3993 & 0.7552 \\
\method{} & 1.00 & 3.445 & 0.2948 & 0.1265 & 0.5777 & 0.2847 & 0.4065 & 0.7579 \\
\method{} & 0.70 & 3.444 & 0.2927 & 0.1253 & 0.4890 & 0.2828 & 0.3957 & 0.7383 \\
\method{} & 0.40 & 3.133 & 0.2871 & 0.1136 & 0.2115 & 0.2717 & 0.3825 & 0.7116 \\
BUDDY (reimpl.) & 1.00 & 3.289 & 0.2984 & 0.1269 & 0.5489 & 0.2699 & 0.3753 & 0.7563 \\
BUDDY (reimpl.) & 0.70 & 2.653 & 0.2589 & 0.0950 & 0.0735 & 0.0961 & 0.1918 & 0.6834 \\
BUDDY (reimpl.) & 0.40 & 1.328 & 0.1554 & 0.0263 & 0.0190 & 0.0665 & 0.1523 & 0.5506 \\
\bottomrule
\end{tabular}%
}
\end{table}

\begin{table}[!htbp]
\centering
\caption{\textbf{Qwen2.5-3B decode efficiency.} TPOT and peak decode VRAM are measured separately from quality, with 128-token prefill and 128 decode-produced tokens. The prefill bootstrap token is excluded from the decode count.}
\label{tab:qwen25-efficiency}
\small
\begin{tabular}{lccc}
\toprule
Method & $\omegaT$ & TPOT (ms/token) & Peak decode VRAM (GiB) \\
\midrule
Vanilla & 1.00 & 3.9012 & 7.77 \\
\method{} & 1.00 & 4.0402 & 7.99 \\
\method{} & 0.70 & 3.4792 & 7.78 \\
\method{} & 0.40 & 2.5997 & 7.58 \\
BUDDY (reimpl.) & 1.00 & 3.9311 & 7.80 \\
BUDDY (reimpl.) & 0.70 & 3.4787 & 7.80 \\
BUDDY (reimpl.) & 0.40 & 2.6533 & 7.80 \\
\bottomrule
\end{tabular}
\end{table}
At requested $\omegaT=0.4$, \method{} has higher scores than BUDDY on all seven reported quality metrics, with TPOT of 2.5997 versus 2.6533 ms/token and peak decode VRAM of 7.58 versus 7.80 GiB. At $\omegaT=0.7$, their TPOTs are nearly identical (3.4792 and 3.4787 ms/token). BUDDY's budget-independent 7.80 GiB peak is consistent with retaining aligned K/V positions; these retained positions also leave later attention histories unshortened. \method{}'s peak decreases from 7.99 to 7.58 GiB across the sweep.

\section{Sensitivity to prompt length}
\label{app:prompt-length}

We evaluate LLaMA-3.1-8B at BS=16 with full-compute prefill
lengths of 128--2048 tokens and 128 decode-produced tokens.
Peak decode VRAM is measured during decoding, including resident
weights and prompt K/V but excluding prefill-only temporaries.
Prompts are left-truncated or attended left-padded to each target
length; padded positions participate in attention, computation, and
cache storage.

\begin{table}[!htbp]
\centering
\caption{Prompt-length sensitivity of \method{} on LLaMA-3.1-8B.
TPOT is in ms/token; peak decode VRAM is in GiB.}
\label{tab:prompt-length}
\small
\begin{tabular}{rccc ccc}
\toprule
& \multicolumn{3}{c}{Decode TPOT}
& \multicolumn{3}{c}{Peak decode VRAM} \\
\cmidrule(lr){2-4}\cmidrule(lr){5-7}
Prompt length
& $\omegaT=1.0$ & $0.7$ & $0.4$
& $\omegaT=1.0$ & $0.7$ & $0.4$ \\
\midrule
128  & 4.6728 & 4.0781 & 3.0392 & 17.05 & 16.88 & 16.62 \\
256  & 5.2226 & 4.6315 & 3.6883 & 17.56 & 17.38 & 17.11 \\
512  & 6.0288 & 5.1887 & 4.0898 & 18.57 & 18.38 & 18.11 \\
1024 & 7.1113 & 6.0759 & 5.0831 & 20.58 & 20.38 & 20.09 \\
2048 & 9.0585 & 7.8505 & 6.8457 & 24.57 & 24.35 & 24.05 \\
\bottomrule
\end{tabular}
\end{table}

Decode TPOT increases because each generated token attends to a longer
full-compute prompt history. Peak decode VRAM increases because the longer
prompt's K/V remains resident throughout decoding. Lowering
$\omegaT$ reduces both quantities at every tested length. In absolute
terms, the TPOT reduction relative to $\omegaT=1.0$ generally becomes
larger for longer prompts: it increases from 0.59 to 1.21 ms/token
between 128 and 2048 prompt tokens at $\omegaT=0.7$, and from 1.63 to
2.21 ms/token at $\omegaT=0.4$. Peak-memory savings grow much less, from 0.17 to 0.22 GiB at
$\omegaT=0.7$ and from 0.43 to 0.52 GiB at $\omegaT=0.4$.
Because prefill always executes all layers, longer prompts add nearly
the same prompt K/V memory at every target. Routing and compact K/V
omission apply only to the fixed 128-token decode.

\section{Variation across training seeds}
\label{app:seeds}

We independently train five additional \method{} models on LLaMA-3.1-8B
with seeds $\{0,1,2,3,4\}$, starting each run from the same unmodified
Meta base checkpoint; the main-results reference checkpoint is not
included in these five runs. All other training and evaluation settings
are held fixed.
Table~\ref{tab:seeds} reports the mean and sample standard
deviation across the five runs at three requested budgets.

\begin{table}[!htbp]
\centering
\caption{Five independent \method{} training runs on LLaMA-3.1-8B.
Entries are mean $\pm$ sample SD.
Prometheus uses the 1--5 scale; other metrics are fractions.}
\label{tab:seeds}
\small
\begin{tabular}{lccc}
\toprule
Metric & $\Omega=1.0$ & $\Omega=0.7$ & $\Omega=0.4$ \\
\midrule
Prometheus-2
& $3.644\pm0.034$ & $3.394\pm0.021$ & $2.036\pm0.030$ \\
ROUGE-L
& $0.2807\pm0.0016$ & $0.2644\pm0.0018$ & $0.1995\pm0.0019$ \\
BLEU-4
& $0.1237\pm0.0019$ & $0.1118\pm0.0016$ & $0.0607\pm0.0014$ \\
GSM8K-Flex
& $0.4238\pm0.0046$ & $0.2177\pm0.0035$ & $0.0206\pm0.0035$ \\
IFEval strict-prompt
& $0.3619\pm0.0058$ & $0.2776\pm0.0060$ & $0.0943\pm0.0037$ \\
IFEval strict-instruction
& $0.4712\pm0.0046$ & $0.3871\pm0.0049$ & $0.1892\pm0.0055$ \\
\bottomrule
\end{tabular}
\end{table}

Across the tested budgets, the sample SD is at most 0.034
for Prometheus, 0.0046 for GSM8K-Flex, and 0.0060 across the IFEval metrics.
The differences between budget settings are
larger than the observed variation across training seeds.
At $\Omega=0.4$, mean GSM8K-Flex accuracy across the five seeds was 2.06\% (SD 0.35 percentage points).

\section{Comparisons with SkipGPT and FlexiDepth}
The experiments in this appendix use LLaMA-3.1-8B.

\subsection{SkipGPT's budget objective}
\label{app:skipgpt-budget}
SkipGPT trains each checkpoint for one sparsity target. Its router objective averages the fraction of skipped modules across tokens in a training micro-batch and penalizes the absolute difference from the sparsity target~\citep{SkipGPT}. If two equal-length requests execute 60\% and 100\% of their modules, their average execution is exactly 80\%, although each request differs from 80\% by 20 percentage points. Tokens within a request may execute different numbers of modules. At inference, the router uses each token's hidden state.

\begin{table}[!htbp]
\centering
\caption{\textbf{Additional 8B Alpaca comparisons.} SkipGPT's target is set during router training and \method{}'s target is supplied per request at inference. FlexiDepth uses a training penalty. Execution counts attention/FFN modules for SkipGPT and whole blocks for the other methods. The counts exclude prefill and the first two full-depth decode positions.}
\label{tab:additional-baselines}
\small
\setlength{\tabcolsep}{4pt}
\begin{tabular}{lrrrrr}
\toprule
Method & Target & Execution & Prometheus-2 & ROUGE-L & BLEU-4 \\
\midrule
SkipGPT ($8\!\times\!3$) & 0.80 & 0.9744 & 3.47 & 0.2705 & 0.1173 \\
FlexiDepth & --- & 0.8918 & 2.541 & 0.2598 & 0.1058 \\
\method{} & 0.80 & 0.8070 & 3.48 & 0.2723 & 0.1184 \\
\method{} & 0.85 & 0.8592 & 3.60 & 0.2735 & 0.1261 \\
\method{} & 0.90 & 0.9057 & 3.63 & 0.2739 & 0.1234 \\
\bottomrule
\end{tabular}
\end{table}

At a target of 0.80, \method{} scores higher than SkipGPT on all three Alpaca metrics in Table~\ref{tab:additional-baselines}. At a target of 0.85, \method{} scores higher than FlexiDepth on all three.

\subsection{SkipGPT retention after LoRA adaptation}
We trained SkipGPT with eight requests per micro-batch and three accumulation steps. Stage 1 trains the router with an 80\% retention target. Stage 2 freezes the router and trains LoRA with the language-model loss. With LoRA disabled, SkipGPT executed 80.75\% of modules. The final model with LoRA enabled executed 97.44\%. LoRA updates change the hidden states seen by the frozen router. \method{} updates the routers and adapters jointly under CE and the request-average budget loss.

\subsection{FlexiDepth execution and generated answers}
\label{app:flexidepth-answers}
FlexiDepth's official 8B configuration places routers and adapters in the last 16 of 32 layers~\citep{FlexiDepthAdaptive}. We use its frozen backbone and training coefficient $\alpha=10^{-4}$. On 1,000 Alpaca responses, 89.18\% of layers execute on average. Forced-full execution bypasses the router and dispatch and takes 4.3356 ms per output token. The routed path takes 4.4199 ms per output token, 1.94\% longer.

For \emph{How do I wrap a present neatly?}, FlexiDepth repeats the steps for taping and cutting the paper. In a recipe answer, it names the dish and then repeats an unrelated phrase about time. Like BUDDY's repeated synonym in Section~\ref{sec:exp-main}, the gift-wrapping answer shares words with the reference without adding another useful step.

\end{document}